\documentclass[lettersize,journal]{IEEEtran}
\usepackage{amsmath,amsfonts}
\usepackage{algorithmic}
\usepackage{algorithm}
\usepackage{array}
\usepackage[caption=false,font=normalsize,labelfont=sf,textfont=sf]{subfig}
\usepackage{textcomp}
\usepackage{stfloats}
\usepackage{url}
\usepackage{verbatim}
\usepackage{graphicx}
\usepackage{cite}
\usepackage{multirow}
\usepackage{multicol}
\usepackage{makecell}
\usepackage{subfig}
\usepackage{bm}
\usepackage[switch]{lineno}
\usepackage{color}

\begin{document}

\title{Spatial-Temporal Knowledge-Embedded Transformer for Video Scene Graph Generation}

\author{Tao Pu, Tianshui Chen, Hefeng Wu, Yongyi Lu, Liang Lin
\thanks{This work was supported in part by National Key R\&D Program of China under Grant No.  2021ZD0111601, National Natural Science Foundation of China (NSFC) under Grant No. 62206060, 62272494, and 61836012, GuangDong Basic and Applied Basic Research Foundation under Grant No. SL2022A04J01626, 2023A1515012845, and 2023A1515011374, and Fundamental Research Funds for the Central Universities, Sun Yat-sen University, under Grant No. 23ptpy111, GuangDong Province Key Laboratory of Information Security Technology. (Corresponding author: Tianshui Chen)}
\thanks{Tao Pu, Hefeng Wu, and Liang Lin are with the School of Computer Science and Engineering, Sun Yat-Sen University, Guangzhou 510006, China. (e-mail: putao3@mail2.sysu.edu.cn, wuhefeng@mail.sysu.edu.cn, linliang@ieee.org)}
\thanks{Tianshui Chen and Yongyi Lu are with the School of Information Engineering, Guangdong University of Technology, Guangzhou 510006, China. (e-mail: tianshuichen@gmail.com, yylu1989@gmail.com).}%
}

% The paper headers
\markboth{IEEE Transactions on Image Processing}%
{Shell \MakeLowercase{\tiextit{et al.}}: A Sample Article Using IEEEtran.cls for IEEE Journals}

%\IEEEpubid{0000--0000/00\$00.00~\copyright~2021 IEEE}
% Remember, if you use this you must call \IEEEpubidadjcol in the second
% column for its text to clear the IEEEpubid mark.

\maketitle

% Paper requirement: https://www.ieeesmc.org/publications/transactions-on-cybernetics/information-for-authors-3/

% up to 250 words
\begin{abstract}
Video scene graph generation (VidSGG) aims to identify objects in visual scenes and infer their relationships for a given video. It requires not only a comprehensive understanding of each object scattered on the whole scene but also a deep dive into their temporal motions and interactions. Inherently, object pairs and their relationships enjoy spatial co-occurrence correlations within each image and temporal consistency/transition correlations across different images, which can serve as prior knowledge to facilitate VidSGG model learning and inference. In this work, we propose a spatial-temporal knowledge-embedded transformer (STKET) that incorporates the prior spatial-temporal knowledge into the multi-head cross-attention mechanism to learn more representative relationship representations. Specifically, we first learn spatial co-occurrence and temporal transition correlations in a statistical manner. Then, we design spatial and temporal knowledge-embedded layers that introduce the multi-head cross-attention mechanism to fully explore the interaction between visual representation and the knowledge to generate spatial- and temporal-embedded representations, respectively. Finally, we aggregate these representations for each subject-object pair to predict the final semantic labels and their relationships. Extensive experiments show that STKET outperforms current competing algorithms by a large margin, e.g., improving the mR@50 by 8.1\%, 4.7\%, and 2.1\% on different settings over current algorithms. 
\end{abstract}


\begin{figure}[!h] 
  \centering
  \includegraphics[width=0.9\linewidth]{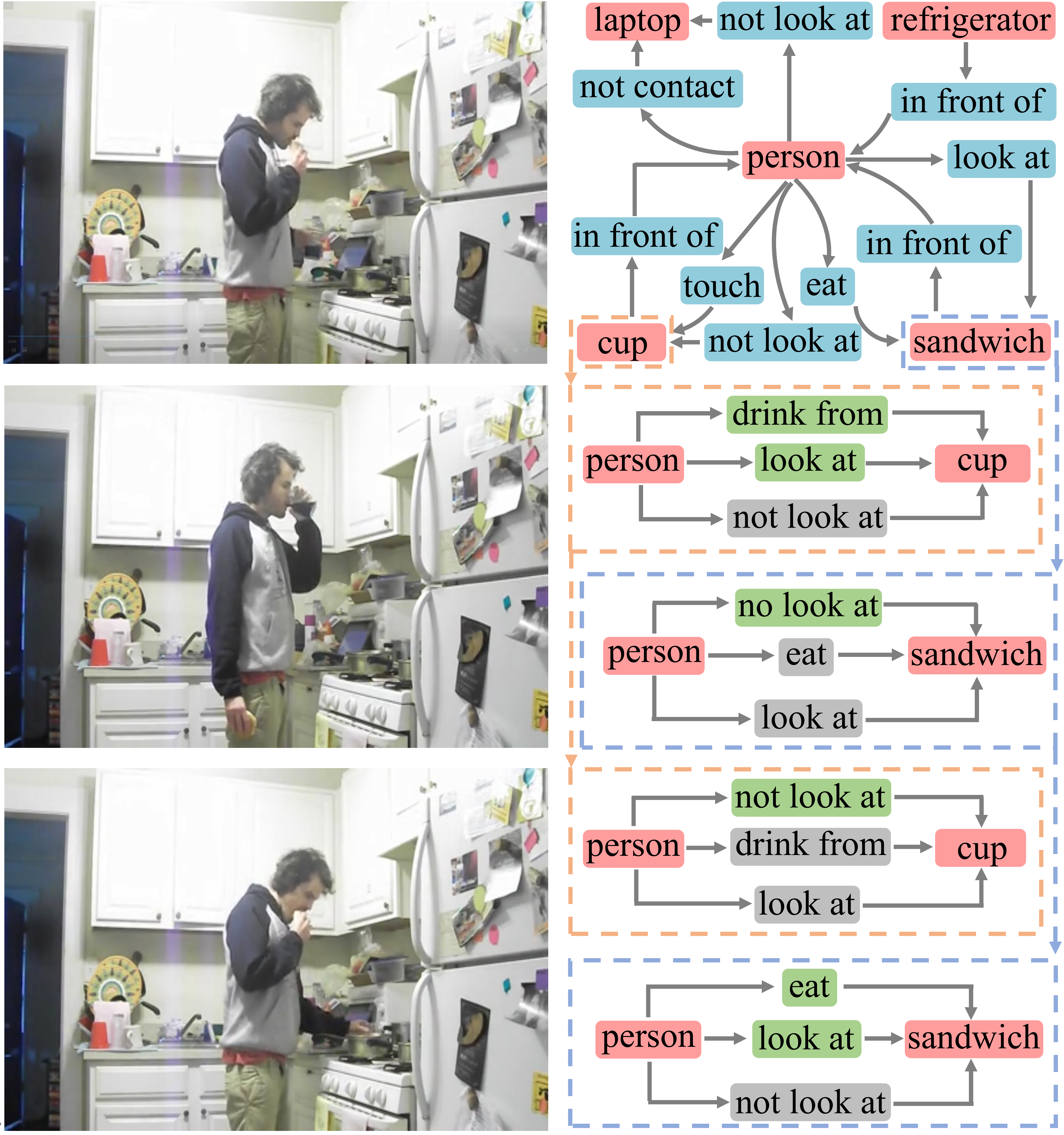}
  \caption{Qualitative results of our proposed STKET. The green color indicates visual relationships that emerge in the current image. The gray color indicates visual relationships that disappeared in the current image.}
  \label{fig:motivation}     
\end{figure}

%%%%%%%%% BODY TEXT
\section{Introduction}
Scene graph generation (SGG) \cite{Xu2017IMP, Ren2020TNNLS-SGG, Tao2022TIP-SGG} aims to depict a visual scene as a structured graph, where nodes correspond to semantic objects and edges refer to the corresponding relationships. It is treated as a promising approach to bridge the significant gap between vision and natural language domains, due to its capacity to represent accurately the semantics of visual contents in the holistic scene. Recently, lots of efforts have been dedicated to demonstrating its efficacy in numerous visual reasoning tasks, including video action recognition \cite{Tu2019TIP-VideoActionRecognition, Han2022TAN, Zhou2023TIP-VideoActionRecognition}, video segmentation \cite{Tang2021TCSVT-VideoSegmentation, Ding2022BCA, Hui2023TPAMI-CMAM}, video captioning \cite{Nishimura2021VPC, Huang2021SemanticTag, Wang2022TIP-VideoCaptioning, Hua2022TIP-VideoCaptioning}, and video question answering \cite{Yang2020BertVideoQA, Zeng2022TIP-VideoQA, Gao2022TIP-VideoQA, Liu2022TIP-VideoQA}. However, the intricate spatial interactions among semantic objects pose a significant challenge to learning visual relationships in static images \cite{Liu2019SGGsurvey, Xu2020SGGsurvey, Chang2021SGGsurvey}. An even more substantial challenge exists in the task of generation scene graphs in videos (VidSGG) \cite{Shang2017VidVRD, Qian2019VRD-GCN, Zheng2022VRDFormer}, as it further requires exploring temporal motions and interactions, hence making VidSGG a formidable yet unresolved task.

Current works \cite{Li2022APT, Cong2021STTran} primarily focus on aggregating object-level visual information from spatial and temporal perspectives to learn relationship representation for VidSGG. In contrast, humans rely on not only visual cues but also accumulated prior knowledge of spatial-temporal correlations to discern ambiguous visual relationships. As illustrated in Figure \ref{fig:motivation}, prior knowledge consists of two aspects. 1) Spatial co-occurrence correlations: the relationship between certain object categories tends toward specific interactions. Given the subject and object of \textit{person} and \textit{sandwich}, their relationship tends to be \textit{eating} or \textit{holding} instead of \textit{wiping}. 2) Temporal consistency/transition correlations: the relationship of a given pair tend to be consistent across continuous video clip or have a high probability of transit to another specific relationship. Given the subject and object of \textit{person} and \textit{cup} with the relationship of \textit{holding} in the current frame, it is likely to keep the \textit{holding} relationship or transit to the relationship of \textit{drinking}. Consequently, integrating these correlations can effectively regularize spatial prediction space within each image and sequential variation space across temporal frames, thereby reducing ambiguous predictions.

In this work, we find that initializing spatial-temporal knowledge embeddings with statistical correlations can better guide model learning spatial-temporal correlations, and the multi-head cross-attention mechanisms can better integrate spatial-temporal correlations with visual information. To this end, we propose a novel spatial-temporal knowledge-embedded transformer (STKET), which introduces multi-head cross-attention layers to incorporate the prior spatial-temporal knowledge for aggregating spatial-temporal contextual information. Specifically, we first initialize the spatial co-occurrence and temporal transition correlations via statistical matrices from the training set and then embed these correlations into learnable spatial-temporal knowledge representations. Then, we design a spatial-knowledge embedded layer to exploit the within-image co-occurrence correlation to guide aggregating spatial contextual information and a temporal-knowledge embedded layer to incorporate cross-image transition correlations to help extract temporal contextual information, which can generate spatial- and temporal-embedded relationship representations, respectively. Finally, we aggregate both spatial- and temporal-embedded relationship representations of each object pair to predict the semantic labels and relationships. Compared with current leading competitors, STKET enjoys two appealing advantages: 1) Integrating these correlations can help to better aggregate spatial and temporal contextual information and thus learn more representative relationship representation to facilitate VidSGG. 2) Incorporating these correlations can effectively regularize relationship prediction, which can evidently reduce the dependencies on training samples and thus dramatically improve the performance, especially for the relationships with limited samples.

The contributions of this work can be summarized in three folds. First, we propose a spatial-temporal knowledge-embedded transformer (STKET) that incorporates spatial co-occurrence and temporal transition correlations to guide aggregating spatial and temporal contextual information, which, on the one hand, facilitates learning more representative relationship representation and, on the other hand, regularizes predication space to reduce the dependencies on training samples. To our knowledge, this is the first attempt to explicitly integrate spatial and temporal knowledge to promote VidSGG. Second, we introduce unified multi-head cross-attention mechanisms to integrate the spatial and temporal correlations via the spatial and temporal knowledge-embedded layers, respectively. Finally, we conduct experiments on the Action Genome dataset \cite{Ji2020AG} to demonstrate the superiority of STKET. It obtains an obvious performance improvement over current state-of-the-art algorithms, especially for the relationships with limited samples, e.g., with the 9.98\%-23.98\% R50 improvement for the top-10 least frequency relationships compared with the previous best-performing algorithm. Codes are available at \textbf{\url{https://github.com/HCPLab-SYSU/STKET}}.

\section{Related Work}

\subsection{Image Scene Graph Generation}
Over the past decade, scene graph generation (SGG) \cite{Zhang2019RelDN, Lin2020GPS-Net, Ren2020HCNet, Lu2021Seq2Seq} has attracted considerable interest across various communities due to its ability to precisely present the semantics of visual contents of complex visual scenes. It aims to identify all objects and their visual relationships within an image, necessitating visual perception and natural language understanding. Hence, much effort has been invested in aligning visual and semantic spaces for relationship representation learning \cite{Lu2016VRD, Li2017MSDN, Yang2018aGCN}. Further research has highlighted the importance of each subject-object pair for inferring ambiguous relationships. Xu et al. \cite{Xu2017IMP} propose to iteratively refine predictions by passing contextual messages, and Zellers et al. \cite{Zellers2018Motif} capture a global context using a bidirectional LSTM. Their impressive performance demonstrates that spatial context is crucial for recognizing visual relationships. Therefore, subsequent research underlines the role of spatial context in generating scene graphs. To this end, many works adopt graph convolutional networks \cite{Kipf2016GCN} or similar architectures to pass messages among different objects. Chen et al. \cite{Chen2019KERN} built a graph that associates detected objects according to these statistical correlations and employs it to learn the context among different objects to regularize prediction. Tang et al. \cite{Tang2019VCTree} design a dynamic tree structure to encode the context among different object regions efficiently. With the impressive progress of transformer \cite{Vaswani2017Transformer}, more and more works propose utilizing this kind of model to learn more representative features from spatial context. Cong et al. \cite{Cong2023Reltr} introduce an encoder-decoder architecture to reason about the visual feature context and visual relationships, using different types of attention mechanisms with coupled subject and object queries. Kundu et al. \cite{Kundu2023IS-GGT} propose contextualized relational reasoning using a two-stage transformer-based architecture for effective reasoning over cluttered, complex semantic structures. Despite impressive progress on static images, leading algorithms may suffer significant performance drops when applied to recognize dynamic visual relationships in videos because it requires an in-depth exploration of temporal consistency/transition correlations across different images.

\subsection{Video Scene Graph Generation}
Building on successfully exploring the spatial context within images, researchers have explored the spatial context and temporal correlation simultaneously in video scene graph generation. With the advent of ImageNet-VidVRD \cite{Shang2017VidVRD}, a benchmark of video visual relation detection, numerous approaches \cite{Qian2019VRD-GCN, Tsai2019GSTEG, Liu2020BeyondShortSnippet, Teng2021TRACE, Zheng2022VRDFormer} have been proposed to employ object-tracking mechanisms to dive into temporal correlations among different image frames. Qian et al. \cite{Qian2019VRD-GCN} propose to use a graph convolution network to pass messages and conduct reasoning in the fully-connected spatial-temporal graphs. Similarly, Tsai et al. \cite{Tsai2019GSTEG} design a gated spatiotemporal energy graph that exploits the statistical dependency between relational entities spatially and temporally. Recently, Teng et al. \cite{Teng2021TRACE} propose a new detect-to-track paradigm by decoupling the context modeling for relation prediction from the complicated low-level entity tracking. Zheng et al. \cite{Zheng2022VRDFormer} propose a unified one-stage model that exploits static queries and recurrent queries to enable efficient object pair tracking with spatiotemporal contexts. However, introducing object-tracking models results in high computational cost and memory consumption and quickly overwhelms the valuable information due to many irrelevant frames, leading to sub-optimal performance.

An alternative line of research tries to address the task of video scene graph generation based on detected object proposals. Compared with tracking-based VidSGG algorithms, this kind of method focuses on modeling the temporal context in the narrow sliding window, avoiding the prediction shift resulting from the inconsistency among tracking proposals. Recently, Cong et al. \cite{Cong2021STTran} propose a strong baseline that adopts a spatial encoder and a temporal decoder to extract implicitly spatial-temporal contexts. This work demonstrates that the temporal correlation is crucial for inferring dynamic visual relationships. Li et al. \cite{Li2022APT} propose a novel anticipatory pre-training paradigm to model the temporal correlation implicitly. Wang et al. \cite{Wang2022TPI} propose to explore temporal continuity by extracting the entire co-occurrence patterns. Kumar et al. \cite{Kumar2023FS-SGG} further propose to spatially and temporally localize subjects and objects connected via an unseen predicate with the help of only a few support set videos sharing the common predicate. These works underscore the critical role of temporal correlation in inferring dynamic visual relationships but overlook the essential prior knowledge of spatial-temporal correlations. Differently, we propose to incorporate spatial-temporal knowledge with multi-head cross-attention layers to learn more representative and discriminative feature representation to facilitate VidSGG. 

Notably, exploiting an object detector to facilitate video understanding is common. Recently, Wang et al. \cite{Wang2020SAOA} propose a novel SAOA framework to introduce the spatial location provided by the detector for egocentric action recognition, which aims to reason the interaction between humans and objects from an egocentric perspective. Although aligning local object features and location proposals to capture the spatial context, the SAOA framework ignores the crucial temporal correlation across continuous frames, which is the main distinction between it and our proposed STKET.

\begin{figure*}[!t] 
  \centering
  \includegraphics[width=0.95\linewidth]{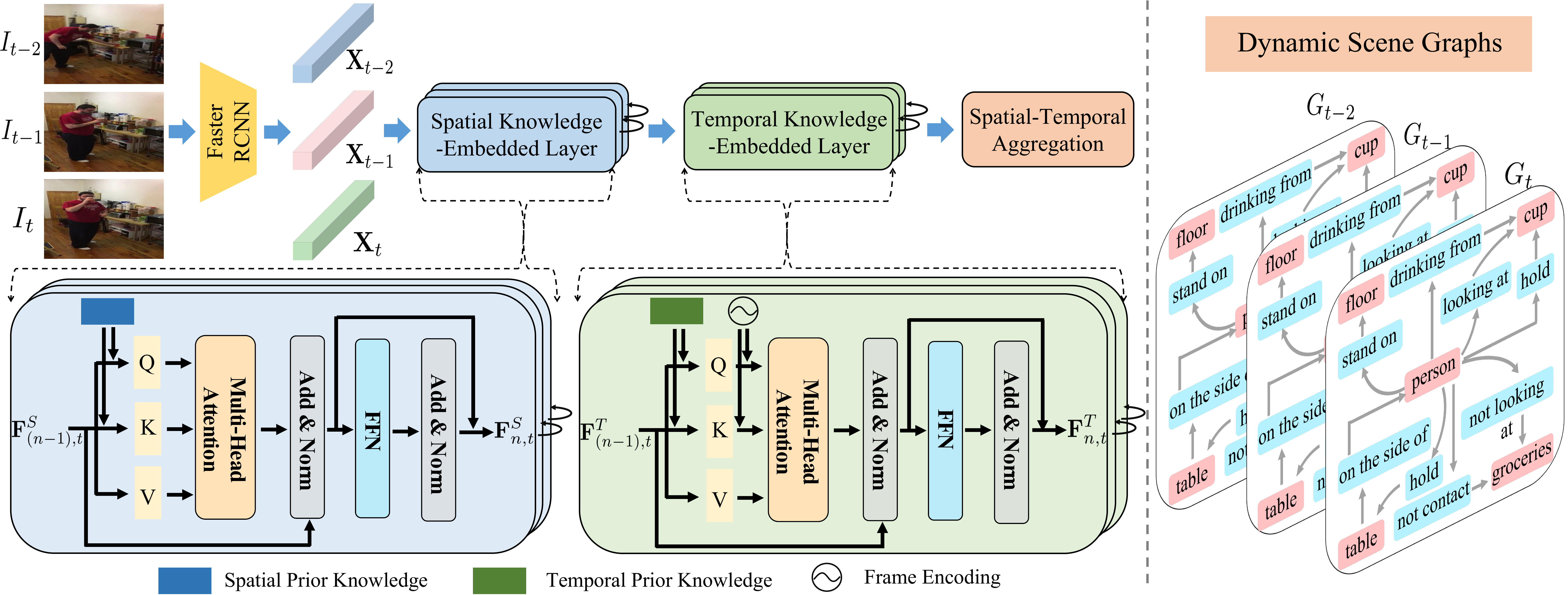}
  \caption{An overall illustration of the proposed STKET framework (left) and its corresponding outputs (right). It first exploits spatial and temporal knowledge-embedded layers to incorporate the spatial co-occurrence and temporal transition correlations into the multi-head cross-attention mechanism to learn spatial- and temporal-embedded representations. Then, it employs a spatial-temporal aggregation module to aggregate these representations for each object pair to predict the final semantic labels and relationships.}
  \label{fig:framework}
  \vspace{0pt}
\end{figure*}

\subsection{Knowledge Representation Learning}
Recent advances in deep learning have allowed neural networks to learn potent representations from raw training data for various tasks \cite{He2016ResNet, Vaswani2017Transformer, Li2022MsKAT, Wang2023Restoreformer++}. However, solely using these vanilla networks may achieve poor performance, especially in weakly supervised learning \cite{Chen2022SST, Pu2022SARB} and domain adaption \cite{Xie2020CD-FER}. To address this challenge, lots of efforts have been made to integrate domain prior knowledge into deep representation learning, resulting in remarkable progress in numerous computer vision tasks, such as few-shot image recognition, facial expression recognition, visual navigation, scene graph generation \cite{Peng2019KTN, Chen2022KGGR, Pu2021AUE-CRL, Chen2022CD-FER, Yang2019ScenePrior, Chen2019KERN}. Specifically, Peng et al. \cite{Peng2019KTN} propose a novel knowledge transfer network that jointly incorporates visual feature learning, knowledge inferring, and classifier learning to fully explore prior knowledge in the few-shot task. Similarly, Chen et al. \cite{Chen2022KGGR} propose a knowledge-guided graph routing framework, which unifies prior knowledge of statistical label correlations with deep neural networks for multi-label few-shot learning. Pu et al. \cite{Pu2021AUE-CRL} introduce the prior knowledge between action unit and facial expression to facilitate facial expression recognition. Yang et al. \cite{Yang2019ScenePrior} incorporate the prior semantic knowledge into a  deep reinforcement learning framework to address the semantic navigation task. Chen et al. \cite{Chen2019KERN} incorporate statistical correlations into deep neural networks to facilitate scene graph generation, in which these statistical correlations between object pairs and their relationships can effectively regularize semantic space and make prediction less ambiguous. However, most of these works merely consider prior spatial knowledge of statistic images, and VidSGG involves spatial-temporal contextual information. In this work, we propose to learn spatial-temporal knowledge and incorporate it into the multi-head cross-attention mechanism to learn more representative relationship representations to facilitate VidSGG.

\section{Method}
\noindent{\textbf{Overview.}} In this section, we first introduce the preliminary of video scene graph generation and then describe the process of extracting spatial co-occurrence and temporal transition correlations from the dataset. Finally, the details of our proposed framework, Spatial-Temporal Knowledge-Embedded Transformer (STKET), are given. 

\subsection{Preliminary}
\noindent{\textbf{Notation.}}  Given a video $V = \{I_1, I_2, ..., I_T\}$, VidSGG aims to detect all visual relationships between objects, presented as a triplet \textless \textit{subject, predicate, object} \textgreater, in each frame to generate a scene graph sequence $G = \{G_1, G_2, ..., G_T\}$, where $G_t$ is the corresponding scene graph of the frame $I_t$. Specifically, define $G_t = \{B_t, O_t, R_t\}$, where $B_t = \{b_t^1, b_t^2, ..., b_t^{N(t)}\}$, $O_t = \{o_t^1, o_t^2, ..., o_t^{N(t)}\}$ and $R_t = \{r_t^1, r_t^2, ..., r_t^{K(t)}\}$ indicate the bounding box set, the object set and the predicate set, respectively. In the frame $I_t$, $N(t)$ is the number of objects, and $K(t)$ is the number of relationships between all objects.

\noindent{\textbf{Relationship Representation.}} For the frame $I_t$, we employ Faster R-CNN \cite{Ren2015FasterRCNN} to provides visual feature representation $\{\mathbf{v}^{1}_{t},...,\mathbf{v}^{N(t)}_{t}\}$, bounding boxes $\{\mathbf{b}^{1}_{t},...,\mathbf{b}^{N(t)}_{t}\}$ and object category distribution $\{\mathbf{d}^{1}_{t},...,\mathbf{d}^{N(t)}_{t}\}$ of object proposals. The relationship representation $\mathbf{x}^{k}_{t}$ of the relationship between the $i$-th and $j$-th object proposals contains visual appearances, spatial information, and semantic embeddings, which can be formulated as
\begin{equation}
\mathbf{x}^{k}_{t} = <f_s(\mathbf{v}^i_t), f_o(\mathbf{v}^j_t), f_u(\varphi(\mathbf{u}^{i,j}_t \oplus f_{box}(\mathbf{b}^i_t, \bm{b}^j_t))), \mathbf{s}^i_j, \mathbf{s}^j_t>, \nonumber
\end{equation}
where $<,>$ is concatenation operation, $\varphi$ is flattening operation and $\oplus$ is element-wise addition. $f_s, f_o$ are implemented respectively by one fully-connected layer which maps a 2048-dimension vector to a 512-dimension vector, and $f_u$ is implemented by one fully-connected layer which maps a 12544-dimension vector to a 512-dimension vector. $\mathbf{u}^{i,j}_t \in \mathbb{R}^{256 \times 7 \times 7}$ is the feature map of the corresponding union box generated by RoIAlign \cite{He2017MaskRCNN} while $f_{box}$ is the function transforming the bounding boxes of subject and object to an entire feature with the same shape as $\mathbf{u}^{i,j}_t$. The semantic embedding vectors $\mathbf{s}^i_t, \mathbf{s}^j_t \in \mathbb{R}^{200}$ are determined by the object category distribution of subject and object. For brevity, we denote all relationship representations in the frame $I_t$ by $\mathbf{X}_t = \{\mathbf{x}^{1}_{t}, ..., \mathbf{x}^{K(t)}_{t} \}$.

\subsection{Spatial-Temporal Knowledge Representation} \label{sec:STKR}
\begin{figure*}[!t] 
  \centering
  \includegraphics[width=0.95\linewidth]{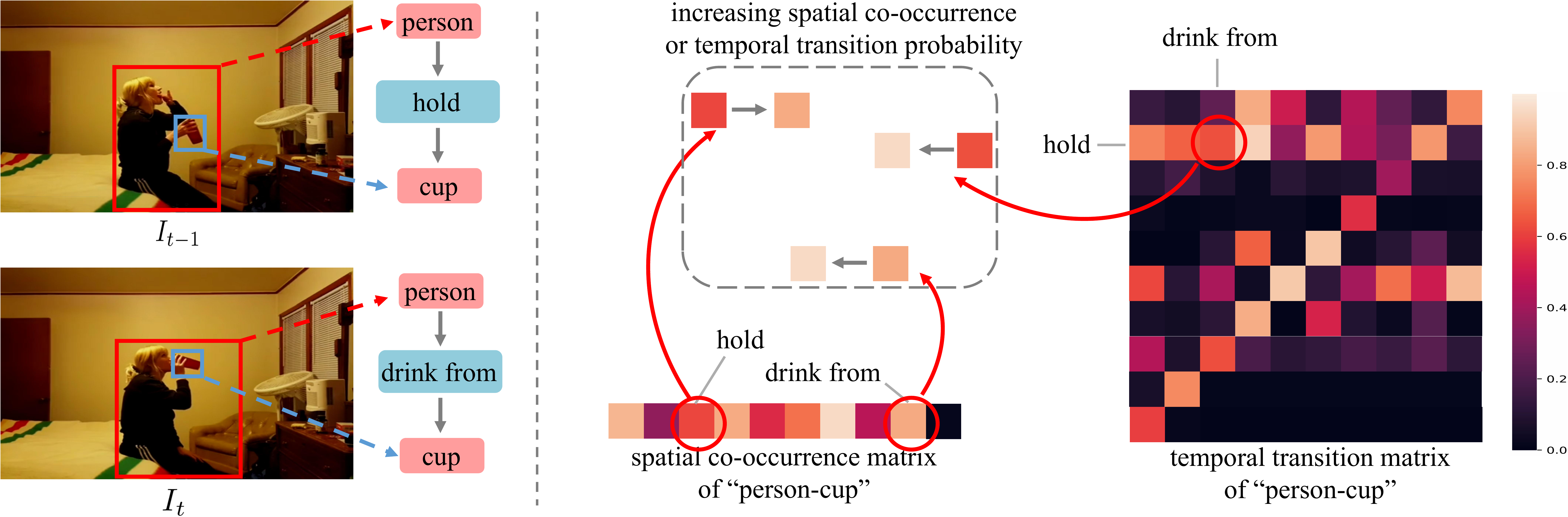}
  \caption{An example of building spatial co-occurrence and temporal transition matrix. Given the video frames and their corresponding relation annotations (left), we first increase the spatial co-occurrence probability of the corresponding predicates (i.e., ``hold" and ``drink from"). Then, for any two consecutive frames, we take the predicate in the previous and current frames as the row and column index separately and increase the temporal transition probability of the corresponding transition pair (i.e., ``hold $\rightarrow$ drink from").}
  \label{fig:prior_knowledge}
  \vspace{0pt}
\end{figure*}

When inferring visual relationships, humans leverage not only visual cues but also accumulated prior knowledge \cite{Vandenbroucke2016Prior}. This approach has been validated on various vision tasks \cite{Lee2018ML-ZSL, Chen2022KGGR, Chen2022CD-FER}. Inspired by this, we propose to distill the prior spatial-temporal knowledge directly from the training set for facilitating the VidSGG task. Specifically, we learn spatial co-occurrence and temporal transition correlations in a statistical manner, as shown in Figure \ref{fig:prior_knowledge}.

\begin{figure}[!t] 
  \centering    
  \subfloat[~]{
  \includegraphics[width=0.9\linewidth]{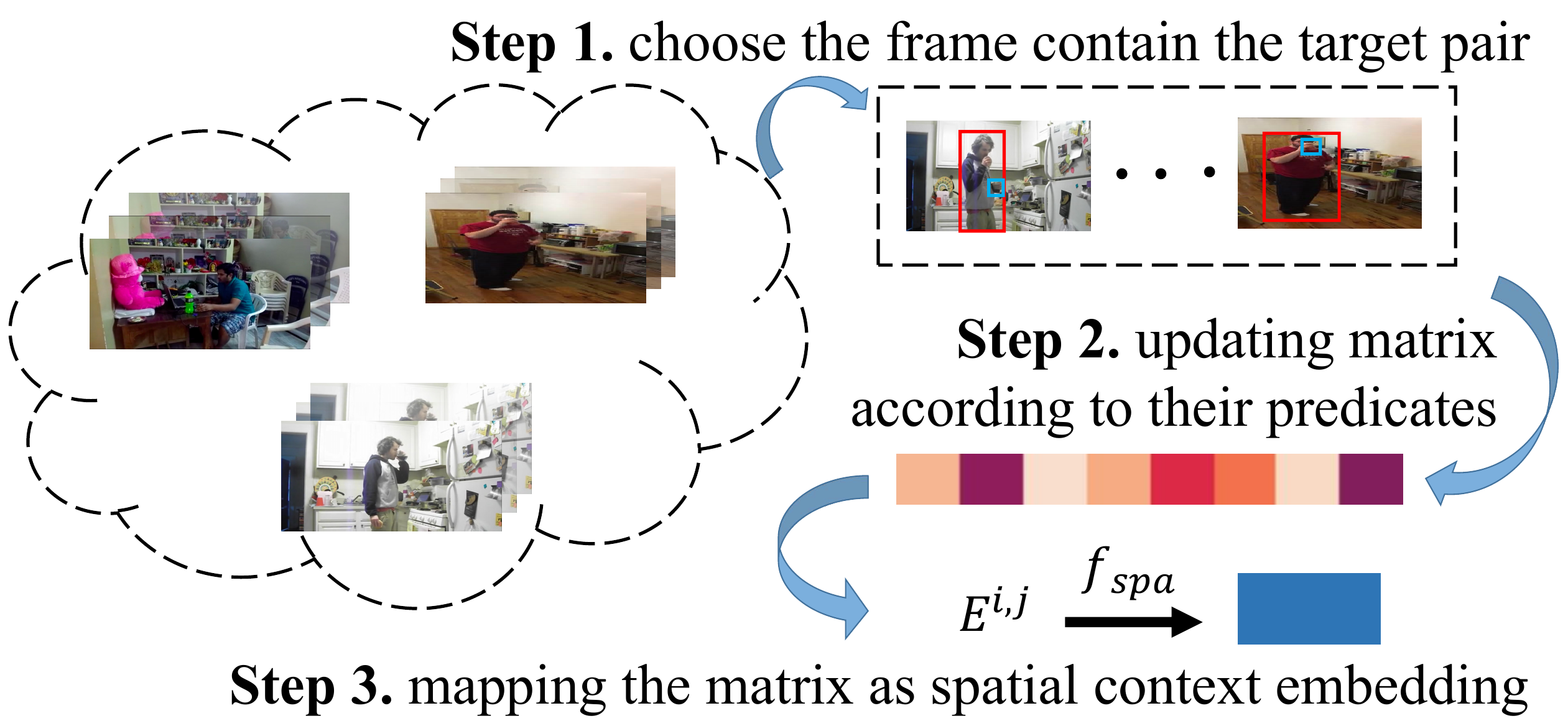}  
  \label{fig:spatial-prior-knowledge}
  }~     
  
  \subfloat[~]{
  \includegraphics[width=0.9\linewidth]{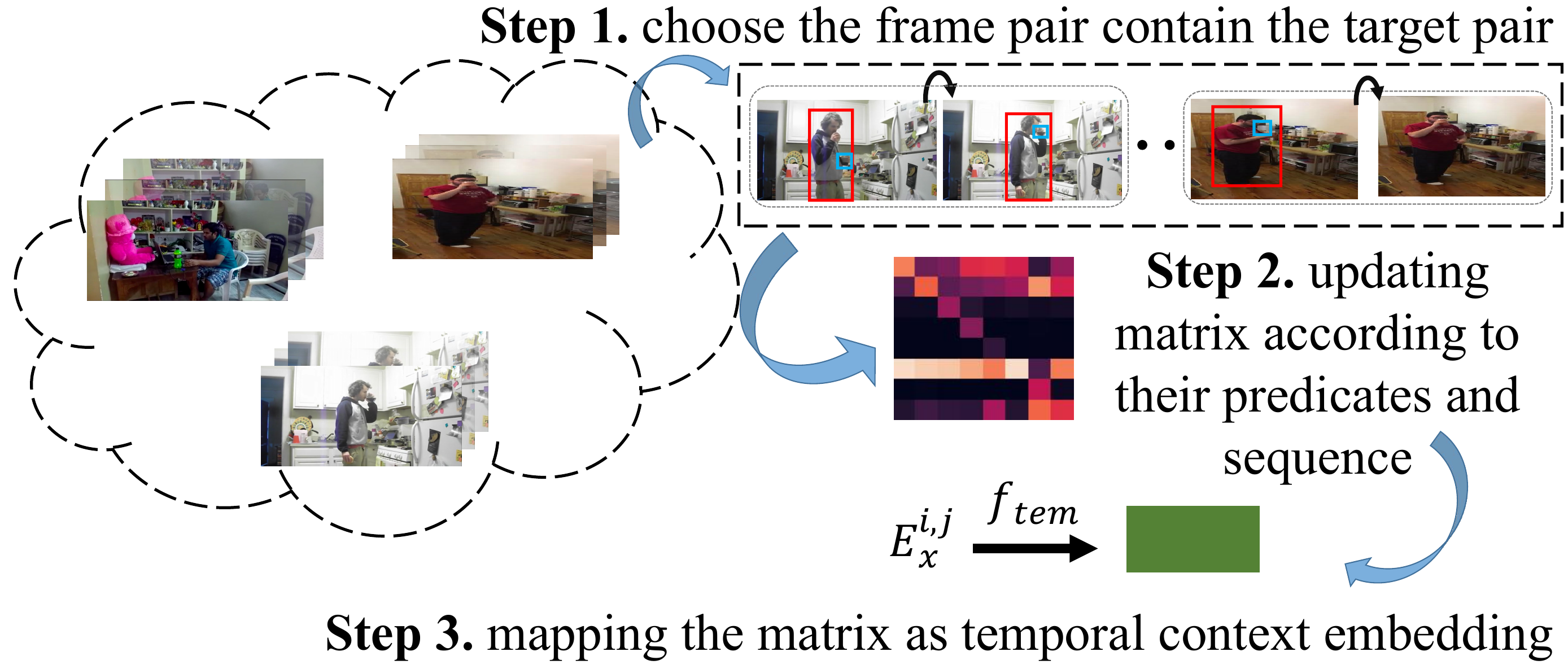} 
  \label{fig:temporal_prior-knowledge}
  }~
  
  \vspace{0pt}
  \caption{Illustration of learning (a) spatial prior knowledge representation and (b) temporal prior knowledge representation.  For better readability, we take the object category pair of \textit{person} and \textit{cup} as an example.}
\end{figure}

\noindent{\textbf{Spatial Prior Knowledge.}} Between different object category pairs, a large difference exists in the spatial co-occurrence correlation of its relationships. To account for this, we construct a spatial co-occurrence matrix $E^{i,j} \in \mathbb{R}^{C}$ for $i$-th object category and $j$-th object category by measuring the frequency of each predicate in the relation set of this pair. $C$ is the total number of types of relationships. For example, let us consider the object category pair of \textit{person} and \textit{cup} (assume their indices are $i$ and $j$, respectively). At each frame, if both objects are present, the total number of co-occurrences of this pair $N^{i,j}$ is incremented by 1. Next, if their predicates contain \textit{holding} (assume the index is $x$), then $e^{i,j}_{x}$ is incremented by 1. Finally, between \textit{person} and \textit{cup}, the spatial co-occurrence probability of each predicate is calculated as:
\begin{equation}
  e^{i,j}_{x} = e^{i,j}_{x} / N^{i,j}, \quad x \in \{1, ..., C\}.
\end{equation}

To distill spatial co-occurrence correlations, the model learns the spatial knowledge embedding based on the spatial co-occurrence matrix. Specifically, the model generates the corresponding spatial knowledge embedding $\mathbf{s}^{k}_{t}$ for $k$-th relationship representation in the frame $I_t$, the generating process can be formulated as
\begin{equation}
  \mathbf{s}^{k}_{t} = f_{spa}( E^{s(k,t),o(k,t)} ),
\end{equation}
where $s(k,t)$ and $o(k,t)$ denote the category of subject and object of $k$-th relationship representation in the frame $I_t$, $f_{spa}(\cdot)$ is implemented by four fully-connected layers which map a $C$-dimension vector to a 1936-dimension vector. For brevity, we denote the corresponding spatial knowledge embeddings for all object pairs in the frame $I_t$ by $\mathbf{S}_{t} = \{ \mathbf{s}^{1}_{t}, ..., \mathbf{s}^{K(t)}_{t} \}$.

Since the distribution of real-world relationships is seriously unbalanced, directly utilizing these spatial knowledge embeddings may degrade the model performance on the less frequent relationships. Thus, we train these embeddings to predict predicates by using the binary cross-entropy loss as the objective function:
\begin{equation}
\mathcal{L}_{spk} = \sum^{T}_{t=1} \sum^{K(t)}_{k=1} \ell(f(\mathbf{s}^{k}_{t}), \mathbf{y}^{k}_{t}),
\end{equation}
where $\ell(\cdot, \cdot)$ denote the binary cross-entropy loss function, $f(\cdot)$ denote the predicate classifier, $\mathbf{y}^{k}_{t}$ denote ground-truth predicate labels of $k$-th relationships in the frame $I_t$.

\noindent{\textbf{Temporal Prior Knowledge.}} In daily life, the interaction between people and objects is characterized by temporal transitions. To identify relationships at different stages, we construct a temporal transition matrix $\hat{E}^{i,j} \in \mathbb{R}^{C \times C}$ for $i$-th object category and $j$-th object category. $C$ is the total number of types of relationships. For instance, let us consider the object category pair of \textit{person} and \textit{cup} as an example (assume the indices are $i$ and $j$, respectively). If the contacting predicate between \textit{person} and \textit{cup} in the previous frame is \textit{holding} (assume the index is $x$) and in the current frame is \textit{drinking} (assume the index is $y$), then $\hat{E}^{i,j}_{x,y}$ is incremented by 1. Finally, between \textit{person} and \textit{cup}, the temporal transition probability of predicates is calculated as:
\begin{equation}
  \hat{e}^{i,j}_{x,y} = \hat{e}^{i,j}_{x,y} / e^{i,j}_{x}, \quad x,y \in \{1, ..., C\}.
\end{equation}

To explore temporal transition correlations, the model learns the corresponding temporal knowledge embedding based on the temporal transition matrix. Specifically, given the predicted object labels and relation labels, the model generates the corresponding temporal knowledge embedding $\mathbf{t}^{k}_{t}$ for $k$-th relationship representation in the frame $I_t$, the generating process can be formulated as
\begin{equation}
  \mathbf{t}^{k}_{t} = f_{tem}( \hat{E}^{s(k,t), o(k,t)}_{r(k,t)} ),
\end{equation}
where $s(k,t)$ and $o(k,t)$ denote the category of subject and object of $k$-th relationship representation in the frame $I_t$, $r(k,t)$ denotes the predicate of $k$-th relationship representation in the frame $I_t$, $f_{tem}(\cdot)$ is implemented by four fully-connected layers which map a $C$-dimension vector to a 1936-dimension vector. Specifically, we utilize the spatial contextualized representation to coarsely predict the relation labels. For brevity, we denote the corresponding temporal knowledge embeddings for all object pairs in the frame $I_t$ by $\mathbf{T}_t = \{\mathbf{t}^{1}_{t}, ..., \mathbf{t}^{K(t)}_{t} \}$.

Since complicated predicate transition exists for different object category pairs, the temporal knowledge embedding may contain inaccurate temporal correlation, resulting in sub-optimal performance. Therefore, we train these embeddings to predict predicates at the next frame by using  the binary cross-entropy loss as the objective function:
\begin{equation}
  \mathcal{L}_{tpk} = \sum^{T-1}_{t=1} \sum^{K(t)}_{i=1} \ell(f(\mathbf{t}^{k}_{t}), \mathbf{y}^{k}_{t+1}),
\end{equation}
where we assume $k$-th relationship representation in the frame $I_{t}$ and frame $I_{t+1}$ belong the same subject-object pair for easily understanding.

\subsection{Knowledge-Embedded Attention Layer} \label{sec:SKEL&TKEL}
Between object pairs and their relationships, there are apparent spatial co-occurrence correlations within each image and strong temporal transition correlations across different images. Thus, we propose incorporating spatial-temporal knowledge into the multi-head cross-attention mechanism to learn spatial- and temporal-embedded representations. 

Spatial knowledge often encapsulates information about positions, distances, and relationships among entities. On the other hand, temporal knowledge concerns the sequence, duration, and intervals between actions. Given their unique properties, treating them separately allows specialized modeling to capture the inherent patterns more accurately. Therefore, we design spatial and temporal knowledge-embedded layers that thoroughly explore the interaction between visual representation and spatial-temporal knowledge.

\noindent{\textbf{Exploring Spatial Context.}} As shown in Figure \ref{fig:framework}, we first employ the spatial knowledge-embedded layers (SKEL) to explore spatial co-occurrence correlations within each image. Specifically, we take all relationship representations of the current frame $I_t$ as the input, i.e., $\mathbf{F}^{S}_{0, t} = \mathbf{X}_t = \{\mathbf{x}^{1}_{t}, ..., \mathbf{x}^{K(t)}_{t} \}$. Then, the SKEL incorporates corresponding spatial knowledge embeddings in the keys and queries to fuse the information from the relationship representation and its corresponding spatial prior. Since we stack $N_s$ identical spatial knowledge-embedded layers, the input of the $n$-th layer is the output of $(n-1)$-th layer. Thus, the queries $\mathbf{Q}$, keys $\mathbf{K}$, values $\mathbf{V}$ and output of the $n$-th SKEL is presented as:
\begin{gather}
  \mathbf{Q} = \mathbf{W}_Q \mathbf{F}^{S}_{(n-1),t} + \mathbf{S}_t,\\
  \mathbf{K} = \mathbf{W}_K \mathbf{F}^{S}_{(n-1),t} + \mathbf{S}_t, \\
  \mathbf{V} = \mathbf{W}_V \mathbf{F}^{S}_{(n-1),t}, \\
  \mathbf{F}^{S}_{n,t} = Att_{spa.}(\mathbf{Q}, \mathbf{K}, \mathbf{V}),
\end{gather}
where the $\mathbf{W}_Q$, $\mathbf{W}_K$ and $\mathbf{W}_V$ denote the linear transformations, the $Att_{spa.}(\cdot)$ denote cross-attention layer in the SKEL. For simplicity, we remove the subscript $n$ and denote the final output of the SKEL as $\mathbf{F}^{S}_{t}$.

\noindent{\textbf{Parsing Temporal Correlation.}} As claimed in prior works \cite{Cong2021STTran, Li2022APT}, the dynamic visual relation can be easily recognized with the given temporal information in the previous frame. Thus, we design the temporal knowledge-embedded layer (TKEL) to explore the temporal correlation in the current frame $I_t$ and the previous frame $I_{t-1}$. At first, we adopt a sliding window over the sequence of spatial contextualized representation $[\mathbf{F}^{S}_{1}, ..., \mathbf{F}^{S}_{T}]$, and the input of the frame $I_t$ in TKEL is presented as:
\begin{equation}
  \mathbf{F}^{T}_{0, t} = [\mathbf{F}^{S}_{t-1}, \mathbf{F}^{S}_{t}], \quad t \in \{2, ..., T\}.
\end{equation}

Then, TKEL incorporates corresponding spatial and temporal knowledge embeddings in the keys and queries to fuse the information from the relationship representation and its corresponding spatial and temporal prior. Since we stack $N_t$ identical temporal knowledge-embedded layers, the input of the $n$-th layer is the output of $(n-1)$-th layer. Thus, the queries $\mathbf{Q}$, keys $\mathbf{K}$, values $\mathbf{V}$ and output of the $n$-th TKEL is presented as:
\begin{gather}
  \mathbf{Q} = \mathbf{W}_Q \mathbf{F}^{T}_{(n-1),t} + [\mathbf{T}_{t-1}, \mathbf{S}_t] + \mathbf{E}_f, \\
  \mathbf{K} = \mathbf{W}_K \mathbf{F}^{T}_{(n-1),t} + [\mathbf{T}_{t-1}, \mathbf{S}_t] + \mathbf{E}_f, \\
  \mathbf{V} = \mathbf{W}_V \mathbf{F}^{T}_{(n-1),t}, \\
  \mathbf{F}^{T}_{n,t} = Att_{tem.}(\mathbf{Q}, \mathbf{K}, \mathbf{V}), 
\end{gather}
where $\mathbf{E}_f$ is the learned frame encoding vector, the $Att_{tem.}(\cdot)$ denote cross-attention layer in TKEL and the $\mathbf{W}_Q$, $\mathbf{W}_K$ and $\mathbf{W}_V$ denote the linear transformations. The intention of adding $[\mathbf{T}_{t-1}, \mathbf{S}_t]$ into the queries and keys is to incorporate the temporal prior about the previous frame and the spatial prior about the current frame. In this way, the TKEL can effectively capture spatiotemporal context in the sliding window.

Considering that the relationships in a frame have various representations in different batches, we choose the earliest representation appearing in the sliding window. For simplicity, we remove the subscript $n$ and denote the final output of TKEL as $\mathbf{F}^{T}_{t}$.

\subsection{Spatial-Temporal Aggregation Module} \label{sec:STA}
As aforementioned, SKEL explores the spatial co-occurrence correlations within each image, and TKEL explores the temporal transition correlations across different images. Though fully exploring the interaction between visual representation and spatial-temporal knowledge, these two layers generate spatial- and temporal-embedded representations, respectively. To explore the long-term context information, we further design the spatial-temporal aggregation (STA) module to aggregate these representations for each object pair to predict the final semantic labels and their relationships. It takes the spatial- and temporal-embedded relationship representations of the identical subject-object pair in different frames as the input. Specifically, we concatenate these representations of the same object pair to generate context representation:
\begin{equation}
   \mathbf{c}^{k}_{t} = cat(\mathbf{f}^{S}_{k,t}, \mathbf{f}^{T}_{k,t}),
\end{equation}
where $\mathbf{f}^{S}_{k,t}$ and $\mathbf{f}^{T}_{k,t}$ denote the spatial- and temporal-embedded representation for the $k$-th relationship in the frame $I_t$. And then, to find the same subject-object pair in different frames, we adopt the predicted object label and the IoU (i.e., intersection over union) to match the same subject-object pairs detected in frames $\{ I_{t-\tau+1}, ..., I_t \}$ (more details in the supplementary material). Thus, the input of $k$-th relationship in the frame $I_t$ in the spatial-temporal aggregation module is presented as
\begin{equation}
   \mathbf{f}^{C}_{k,t} = [\mathbf{c}^{k}_{t-\tau+1}, ..., \mathbf{c}^{k}_{t}],
\end{equation}
where we assume $k$-th relationship representation in frame $\{ I_{t-\tau+1}, ..., I_t \}$ belong same subject-object pair for easily understanding. And its corresponding output of the spatial-temporal aggregation module is presented as:
\begin{gather}
    \mathbf{Q} = \mathbf{W}_Q \mathbf{f}^C_{k,t} + \mathbf{E}^{'}_{f}, \\
    \mathbf{K} = \mathbf{W}_K \mathbf{f}^C_{k,t} + \mathbf{E}^{'}_{f}, \\
    \mathbf{V} = \mathbf{W}_V \mathbf{f}^C_{k,t}, \\
    \hat{\mathbf{f}}^C_{k,t} = Att(\mathbf{Q}, \mathbf{K}, \mathbf{V}),
\end{gather}
where $\mathbf{E}^{'}_f$ is the learned frame encoding vector, the $Att(\cdot)$ denote the self-attention layer in the spatial-temporal aggregation module, and the $\mathbf{W}_Q$, $\mathbf{W}_K$ and $\mathbf{W}_V$ denote the linear transformations. Considering the relationships in a frame have various representations in different batches, we choose the earliest representation appearing in the sliding window.

\subsection{Loss Function}
In the real world, there exist different kinds of relationships between two objects at the same time. Thus, we introduce the binary cross-entropy loss as the objective function for predicate classification as follows:
\begin{equation}
  \ell(r) = \sum_{p\in\mathcal{P}^{+}} log(\phi(r,p)) + \sum_{q\in\mathcal{P}^{-}} log(1-\phi(r,q)).
  \label{eq:bce-loss}
\end{equation}
For a subject-object pair $r$, $\mathcal{P}^{+}$ are the annotated predicates, while $\mathcal{P}^{-}$ is the set of the predicates not in the annotation. $\phi(r,p)$ indicates the computed confidence score of the $p$-th predicate. 

During training, we adopt the binary cross-entropy loss as the objective function for supervising SKEL, TKEL, and STA and denote the corresponding loss by $\ell^{s}(r)$, $\ell^{t}(r)$  and $\ell^{c}(r)$ respectively. Therefore, the final classification loss is defined as summing the three losses over all samples, formulated as
\begin{equation}
 \begin{aligned}
  \mathcal{L}_{cls} &= \sum^{T}_{t=1} \sum^{K(t)}_{r=1}{ [ \ell^{s}(r) + \ell^{t}(r) + \ell^{c}(r) ] }.
 \end{aligned}
\label{eq:cls-loss}
\end{equation}
Therefore, the total objective is formulated as:
\begin{equation}
  \mathcal{L} = \mathcal{L}_{cls} + \mathcal{L}_{spk} + \mathcal{L}_{tpk}.
\label{eq:total-loss}
\end{equation}
In all experiments, we set the loss weights of the three losses to be equal, primarily to ensure that each loss component has an equal contribution to the overall optimization objective.

\begin{table*}[!t]
  \setlength\tabcolsep{5pt}
  \centering
  \caption{Comparison with state-of-the-art scene graph generation methods on Action Genome \cite{Ji2020AG}.}
  \begin{tabular}{c|ccc|ccc|ccc|c}
  \hline
  \hline
  \centering \multirow{2}*{Method} & \multicolumn{3}{c}{Pred Cls} & \multicolumn{3}{c}{SG Cls} & \multicolumn{3}{c|}{SG Gen} & \multirow{2}*{Average Recall} \\
  \cline{2-4} \cline{5-7} \cline{8-10}
  \centering & R@10 & R@20 & R@50 & R@10 & R@20 & R@50 & R@10 & R@20 & R@50 & \\
  \hline
  \hline
  \centering KERN \cite{Chen2019KERN} & 73.7 & 92.3 & 98.8 & 50.5 & 62.4 & 66.3 & 23.3 & 35.2 & 47.0 & 61.1 \\
  \centering VCTREE \cite{Tang2019VCTree} & 75.5 & 92.9 & 99.3 & 52.4 & 62.0 & 65.1 & 23.9 & 35.3 & 46.8 & 61.5 \\
  \centering ReIDN \cite{Zhang2019RelDN} & 75.7 & 93.0 & 99.0 & 52.9 & 62.4 & 65.1 & 24.1 & 35.4 & 46.8 & 61.6 \\
  \centering GPS-Net \cite{Lin2020GPS-Net} & 76.2 & 93.6 & 99.5 & 53.6 & 63.3 & 66.0 & 24.4 & 35.7 & 47.3 & 62.2 \\
  \centering STTran \cite{Cong2021STTran} & 77.9 & 94.2 & 99.1 & 54.0 & 63.7 & 66.4 & 24.6 & 36.2 & 48.8 & 62.8 \\
  \centering TPI \cite{Wang2022TPI} & 78.2 & 94.6 & 99.1 & 54.4 & 64.3 & 67.1 & 25.2 & 36.9 & 49.3 & 63.2 \\
  \centering APT \cite{Li2022APT} & 78.5 & 95.1 & 99.2 & 55.1 & 65.1 & \textbf{68.7} & 25.7 & 37.9 & 50.1 & 63.9 \\
  \hline
  \centering Ours & \textbf{82.6} & \textbf{96.3} & \textbf{99.9} & \textbf{57.1} & \textbf{65.3} & 67.1 & \textbf{27.9} & \textbf{38.8} & \textbf{50.6} & \textbf{65.1} \\
  \hline
  \hline
  \end{tabular}
  \vspace{0pt}
  \label{tab:no-constraint-result}
  \vspace{0pt}
\end{table*}

\begin{table*}[!t]
  \setlength\tabcolsep{5pt}
  \centering
  \caption{Comparison with state-of-the-art scene graph generation methods on Action Genome \cite{Ji2020AG}.}
  \begin{tabular}{c|ccc|ccc|ccc|c}
  \hline
  \hline
  \centering \multirow{2}*{Method} & \multicolumn{3}{c}{Pred Cls} & \multicolumn{3}{c}{SG Cls} & \multicolumn{3}{c|}{SG Gen} & \multirow{2}*{Mean} \\
  \cline{2-4} \cline{5-7} \cline{8-10}
  \centering & mR@10 & mR@20 & mR@50 & mR@10 & mR@20 & mR@50 & mR@10 & mR@20 & mR@50 & \\
  \hline
  \hline
  \centering KERN \cite{Chen2019KERN} & 12.3 & 14.7 & 18.4 & 8.1 & 9.3 & 10.3 & 4.4 & 5.3 & 8.2 & 10.1 \\
  \centering STTran \cite{Cong2021STTran} & 15.4 & 19.6 & 21.2 & 10.7 & 13.1 & 14.0 & 5.0 & 7.3 & 10.2 & 12.9 \\
  \centering TPI \cite{Wang2022TPI} & 16.3 & 20.8 & 22.4 & 11.1 & 13.5 & 14.2 & 5.3 & 6.5 & 9.7 & 13.3 \\
  \centering APT \cite{Li2022APT} & 17.2 & 22.5 & 25.6 & 11.5 & 13.8 & 14.6 & 5.8 & 6.9 & 10.1 & 14.2 \\
  \hline
  \centering Ours  & \textbf{20.1} & \textbf{27.3} & \textbf{33.7} & \textbf{15.4} & \textbf{17.2} & \textbf{19.3} & \textbf{7.4} & \textbf{9.7} & \textbf{12.2} & \textbf{18.0} \\
  \hline
  \hline
  \end{tabular}
  \vspace{0pt}
  \label{tab:no-constraint-meanRecall}
  \vspace{0pt}
\end{table*}

\section{Experiments}

\subsection{Experiment Setting}

\noindent {\textbf{Dataset}} As the most widely used dataset for evaluating video scene graph generation, the Action Genome \cite{Ji2020AG} contains 476,229 bounding boxes of 35 object classes (without the person) and 1,715,568 instances of 26 relationship classes annotated for 234, 253 frames. These 26 relationships are subdivided into three different types: (1) \textit{attention}, (2) \textit{spatial}, and (3) \textit{contact} whose number of categories are 3, 6, and 17, respectively. In all experiments, we use the same training and testing split in previous works \cite{Cong2021STTran, Li2022APT}.

\noindent {\textbf{Task}} Following prior arts \cite{Chen2019KERN, Teng2021TRACE, Cong2021STTran}, we evaluate our proposed method and other state-of-the-art methods under three kinds of experiment setups: \textbf{Predicate Classification} (\textit{Pred Cls}): predict the predicates of object pairs with given ground truth bounding boxes and category labels. \textbf{Scene Graph Classification} (\textit{SG Cls}): predict both the predicates and the category labels of objects with given ground-truth bounding boxes. \textbf{Scene Graph Generation} (\textit{SG Gen}): simultaneously detects objects appearing in the image and predicts the predicates of each object pair. In the \textit{SG Gen}, an object bounding box is considered to be correctly detected only if the predicted object bounding box has at least 0.5 IoU (Intersection over Union) overlap with the ground-truth object bounding box. We use \textbf{No Constraint} strategy of generating a scene graph to evaluate the models. This strategy allows each subject-object pair to have multiple predicates simultaneously. 

\noindent {\textbf{Evaluation Metric}}  All tasks are evaluated with the Recall@K (short as R@$K$) metric ($K$ = [10, 20, 50]), which measures the ratio of correct instances among the top-$K$ predicted instances with the highest confidence. We also report the results by using the mean Recall@$K$ (short as mR@$K$) metric that averages R@$K$ over all relationships. 

\noindent {\textbf{Training Details }} Following previous works \cite{Cong2021STTran, Li2022APT}, we adopt the Faster RCNN \cite{Ren2015FasterRCNN} with a ResNet-101 \cite{He2016ResNet} backbone as the object detector. We first train the detector on the training set of Action Genome \cite{Ji2020AG} and get 24.6 mAP at 0.5 IoU with COCO metrics. The detector is applied to all baselines for fair comparisons. The parameters of the object detector (the object classifier excluded) are fixed when training scene graph generation models. Per-class non-maximal suppression at 0.4 IoU (Intersection over Union) is applied to reduce region proposals provided by RPN.

We use an AdamW \cite{Loshchilov2017Adam} optimizer with initial learning rate $2e^{-5}$ and batch size 1 to train our model. Moreover, gradient clipping is applied with a maximal norm of 5. All experiments are implemented by PyTorch \cite{Paszke2019Pytorch}. In the spatial-temporal aggregation module, we set the size of the sliding window $\tau$ to 4. In KEAL, we stack two identical spatial knowledge-embedded layers to explore spatial co-occurrence correlations and then stack two identical temporal knowledge-embedded layers to temporal transition correlations. The cross-attention and self-attention layers in our proposed framework have 8 heads with $d = 1936$ and $dropout = 0.1$. In SKEL and TKEL, the $1936$-dimension input is projected to $2048$-dimension by the feed-forward network, then projected to $1936$-dimension again after ReLU activation. In STA, the $3872$-dimension input is projected to $1936$-dimension.

\noindent\textbf{Frame Encoding} Unlike the SKEL, the TKEL adopts a sliding window over the sequence of spatial contextualized representation as its input. Therefore, we need to introduce the learned frame encoding $E_f$ into the TKLE to help it fully understand the temporal dependencies among different frames. Specifically, we construct the frame encodings $E_f$ using learned embedding parameters. Specifically, $E_f = [e_1, e_2]$, where $e_1$ and $e_2 \in \mathbb{R}^{1936}$ are the learned vectors. Similarly, we also introduce frame encodings $E^{'}_f$ in STA, where $E^{'}_f = [e^{'}_1, ..., e^{'}_{\tau}]$,  and each encoding is the learned vectors with a length of $3872$.

\begin{figure*}[!t] 
  \centering
  \includegraphics[width=0.90\linewidth]{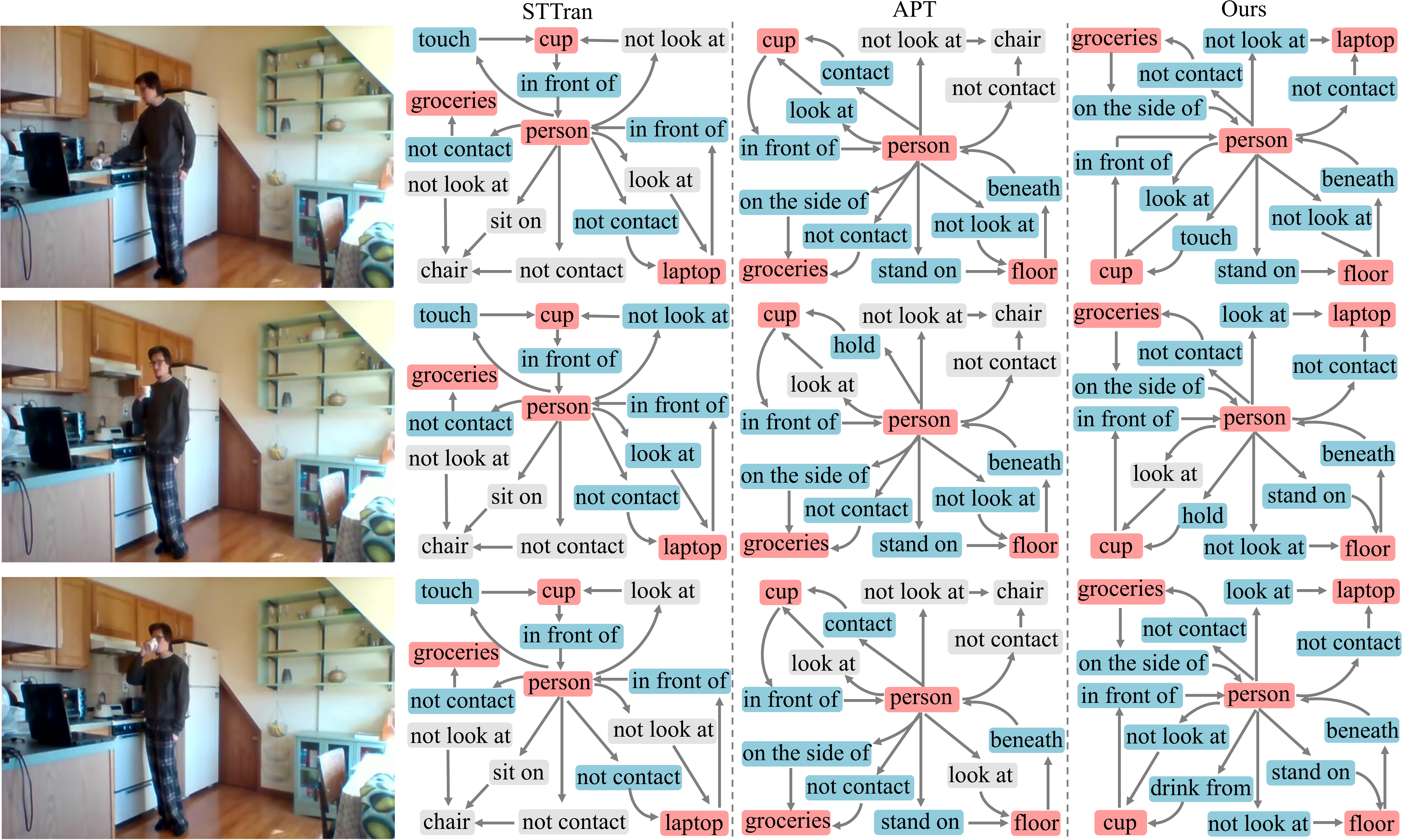}
  \caption{Qualitative results in \textit{SG Gen} task with top-10 confident predictions. The blue and red colors indicate correct relationships and objects, respectively. The gray colors indicate wrong relationships and objects. For fairness, the same object detector is used for all methods.}
  \label{fig:qualitative_result}     
  \vspace{0pt}
\end{figure*}

\noindent\textbf{Pair Tracking} As described in the manuscript, STA takes the spatial- and temporal-embedded representation of the same subject-object pair in different frames as input. We first use the predicted object labels to distinguish different pairs to match the subject-object pairs detected in different frames. If multiple entities of the same category exist, we calculate the intersection over union (IoU) between the two objects across different images to match the subject-object pair. Specifically, we compute the IoU between the bounding box of the target object in the previous frame and that of each object with the same category label in the current frame. If the IoU is higher than 0.8, we consider them to be the same object. We choose the one with the highest IoU if there are multiple candidates.

\subsection{Comparison with State-Of-The-Art Methods} \label{sec:performance}
To evaluate the effectiveness of STKET, we compare it with existing state-of-the-art VidSGG algorithms, including STTran \cite{Cong2021STTran}, TPI \cite{Wang2022TPI}, and APT \cite{Li2022APT}. Previous works \cite{Cong2021STTran} also adapt image-based SGG algorithms to address the VidSGG task by applying the inference process to each frame. We also follow these works to include the image-based SGG algorithms for more comprehensive evaluations and comparisons, including VCTREE \cite{Tang2019VCTree}, KERN \cite{Chen2019KERN}, ReIDN \cite{Zhang2019RelDN}, GPS-Net \cite{Lin2020GPS-Net}. 

We first present the comparison on R@K in Table \ref{tab:no-constraint-result}. As shown, recent video-based algorithms (e.g., STTran, TPI, and APT) obtain quite a marginal improvement over the image-based SGG algorithms as they further introduce temporal contextual information. By introducing spatial-temporal prior knowledge to guide aggregating spatial and temporal contextual information, the proposed STKET framework consistently outperforms in nearly all settings. For example, it improves the R@10 from 78.5\% to 82.6\%, 55.1\% to 57.1\%, and 25.7\% to 27.9\% on the three tasks, with the improvement of 4.1\%, 2.0\% and 2.2\%, respectively. It also obtains similar improvement on R@20 and R@50 metrics. 

\begin{figure}[!h] 
  \centering    
  \subfloat[~]{
    \includegraphics[width=0.9\linewidth]{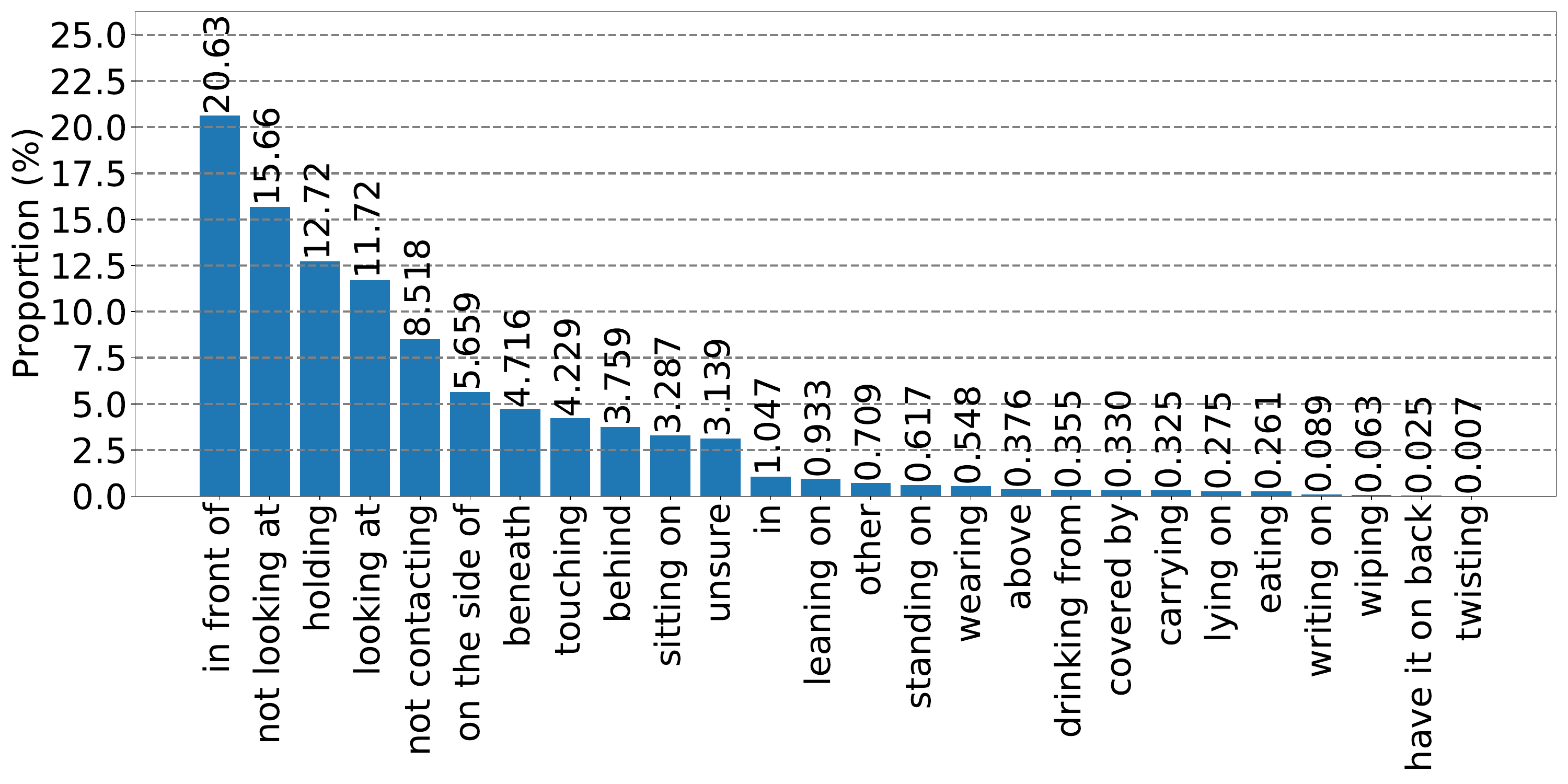}  
    \label{fig:relation_distribution}   
  }~     
  
  \subfloat[~]{
    \includegraphics[width=0.9\linewidth]{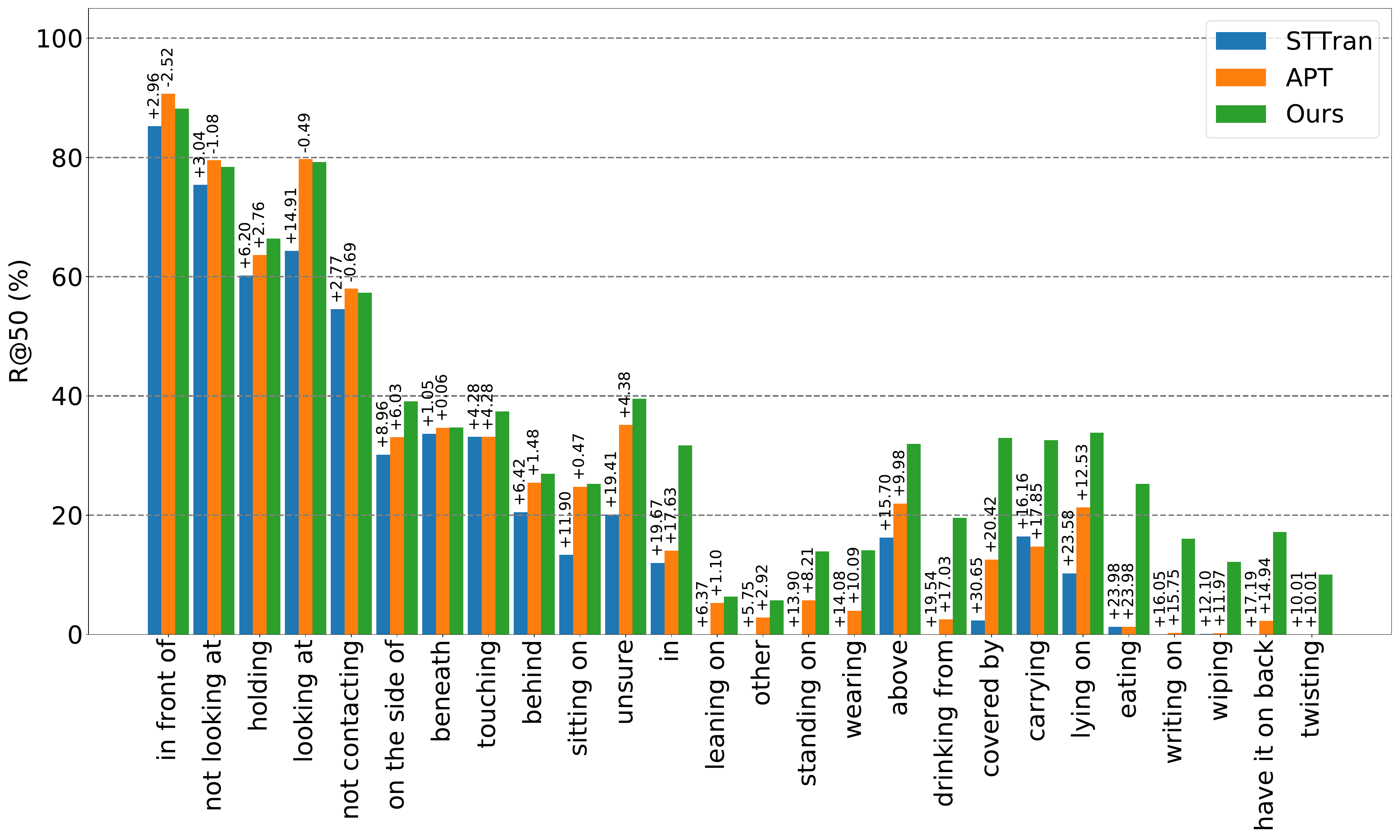}     
    \label{fig:recall_comparison}   
  }~
  \caption{(a) The distribution of different relationships on Action Genome \cite{Ji2020AG}. (b) The R@50 results in \textit{Pred Cls} task of our method, STTran and APT on Action Genome \cite{Ji2020AG}.}  
  \vspace{0pt}
\end{figure}

Considering the mR@K metric offers a better performance measure under uneven distribution \cite{Chen2019KERN}, we also present the comparison results on this metric. As presented in Table \ref{tab:no-constraint-meanRecall}, STKET obtains even more significant improvement, enhancing mR@50 by 8.1\%, 4.7\%, and 2.1\% compared with the current best-performing APT algorithm. To highlight the necessity of introducing the mR@K metric, we present the distribution across different relationships on the AG dataset in Figure \ref{fig:relation_distribution}. As depicted, the distribution of relationships is exceedingly long-tailed, in which the top-10 most frequent relationships occupy 90.9\% samples while the top-10 least frequency relationships merely occupy 2.1\%. In Figure \ref{fig:recall_comparison}, we further provide the detailed performance of each relationship to understand the performance variance across different relationships better. Evidently, current algorithms like STTran and APT deliver competitive performance for relationships with abundant training samples (e.g., ``in front of", ``not looking at") but falter significantly for relationships with limited training samples (e.g., ``wiping", ``twisting"). In contrast, for the top-10 least frequent relationships, STKET improves the R@50 from 9.98\% to 23.98\% compared to the second-best APT and from 12.10\% to 30.65\% compared to the baseline STTran. These comparisons demonstrate STKET can effectively regularize VidSGG training and thus reduce the dependencies on training samples by explicitly incorporating spatial-temporal prior knowledge. 

\begin{figure}[!t] 
  \centering    
  \subfloat{
  \includegraphics[width=0.45\linewidth]{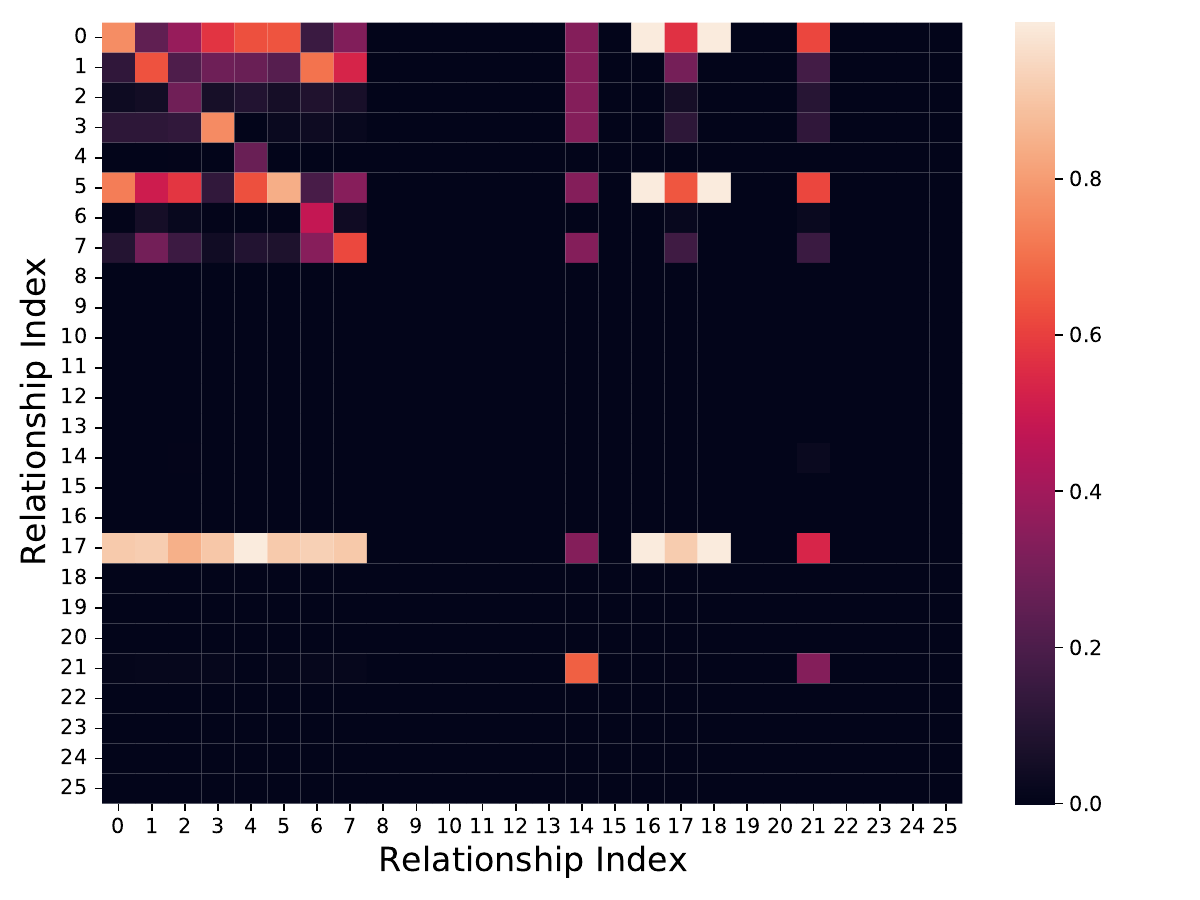}  
  }~     
  \subfloat{
  \includegraphics[width=0.45\linewidth]{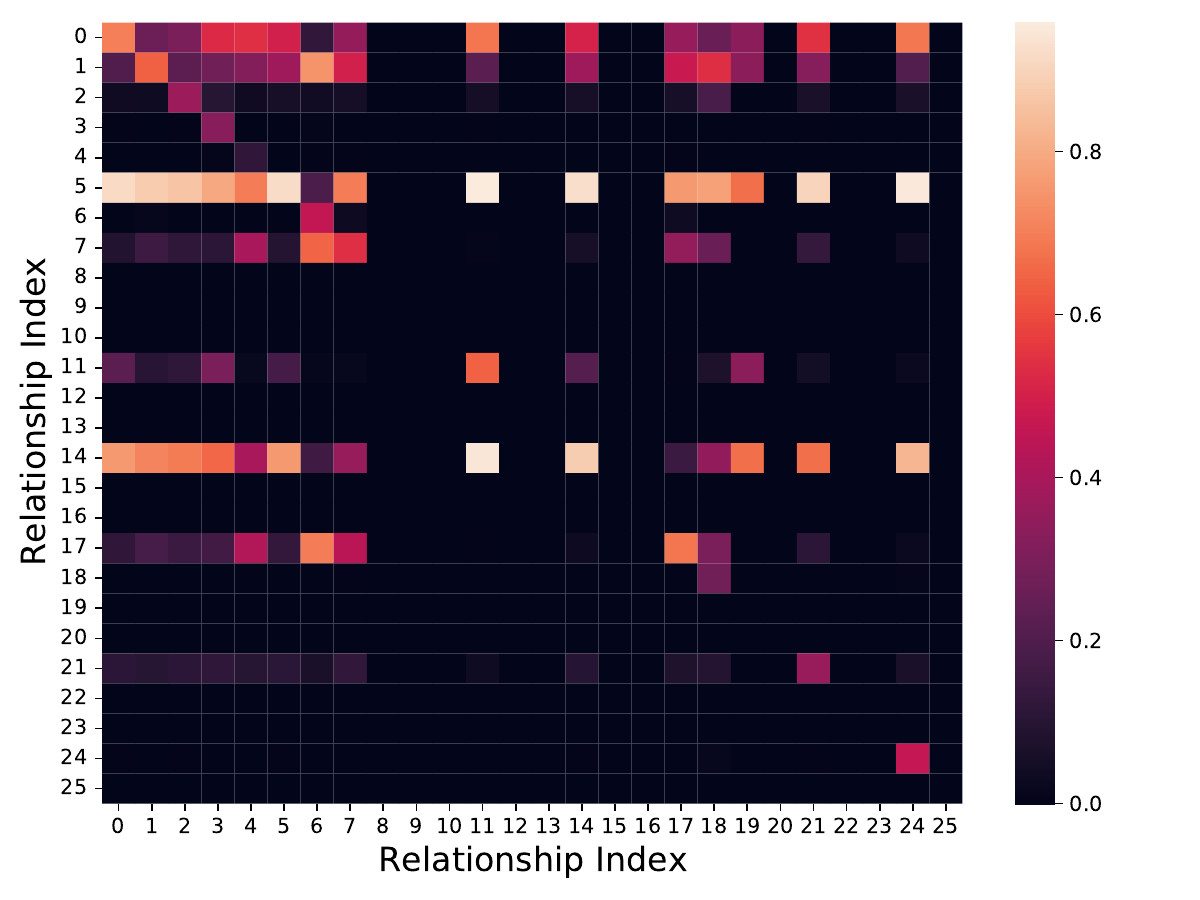} 
  }~
  
  \subfloat{
  \includegraphics[width=0.95\linewidth]{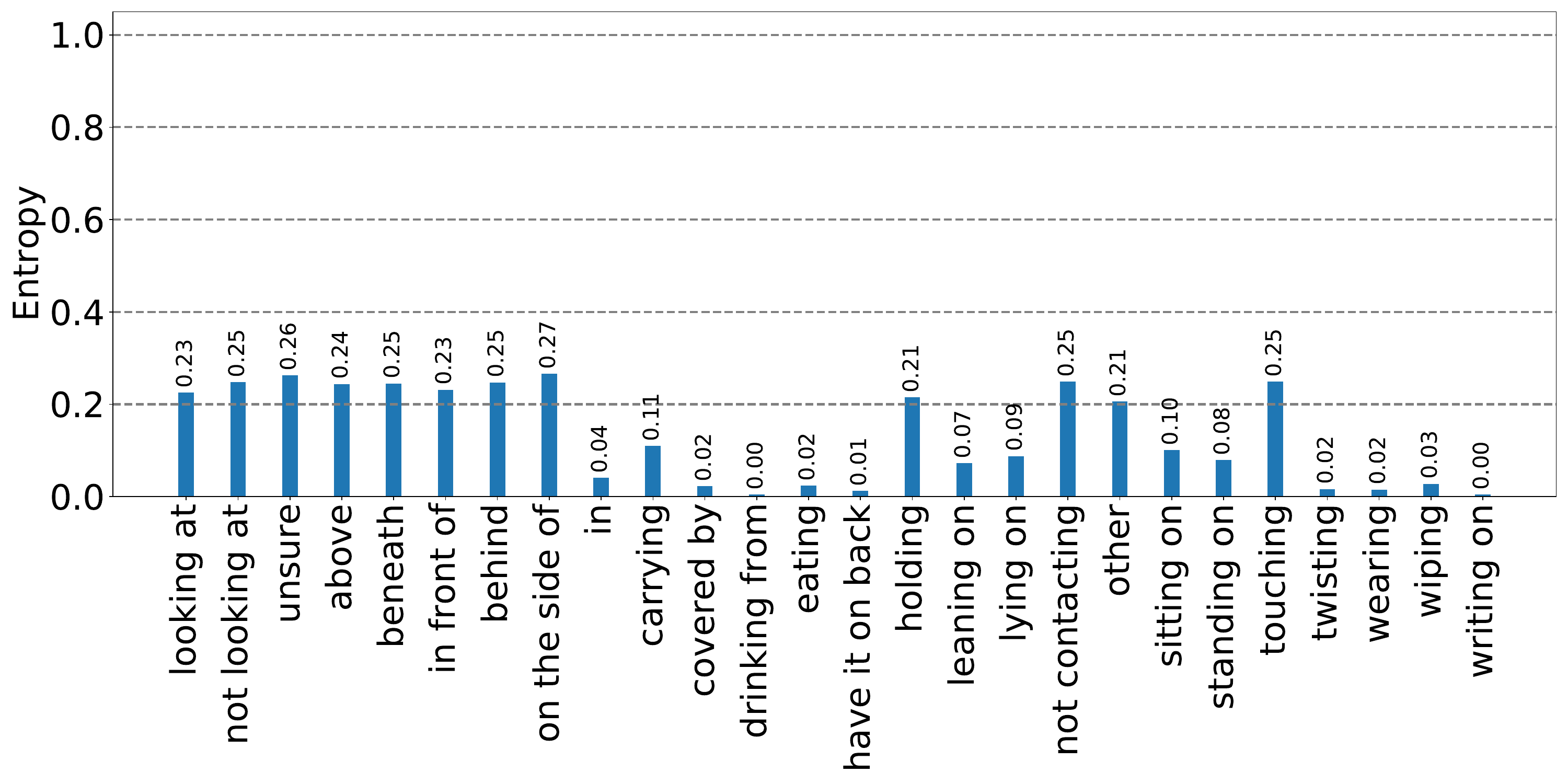}
  }~
  \vspace{0pt}
  \caption{Top: Relationship transition matrix of ``man-tv" (left) and ``man-cup" (right). For brevity, we print the relation index instead of the relation label (see the corresponding labels below figure, i.e., 0 denotes ``looking at" and 25 denotes ``writing on"). Bottom: The entropy of spatial-temporal knowledge across different predicates.} 
  \label{fig:entropy}
\end{figure}

In Figure \ref{fig:qualitative_result}, we visualize the qualitative results of our method and current leading VidSGG methods (i.e., STTran and APT). The results show that existing state-of-the-art VidSGG algorithms tend to predict numerous false-positive relationships, while our method successfully predicts nearly all relationships with high accuracy. This highlights the strength of integrating spatial-temporal knowledge in discerning dynamic relationships, especially within intricate interactions. For instance, across all frames, our STKET accurately identifies the attentional relationship among \textit{person}, \textit{cup}, and \textit{laptop}, where current leading VidSGG algorithms stumble. It is also noteworthy that our method not only performs well on high-frequency relationships but also on low-frequency relationships. For instance, in the last frame, our method correctly predicts the predicate of ``drink from" whose samples merely occupy 0.355\% of the total samples, while other methods miss it. This success is largely due to our method's explicit incorporation of spatial-temporal correlations, which helps to lessen the dependency on training samples significantly.

\subsection{Ablative Study}
In this section, we conduct comprehensive experiments to analyze the actual contribution of each crucial component. Here, we mainly present the R@10, R@20, mR@10, and mR@20 on \textit{Pred Cls} and \textit{SG Cls} as they can better describe the performance.

\subsubsection{Analysis of STKET} As aforementioned, STKET integrates spatial-temporal knowledge into the multi-head cross-attention mechanism to aggregate spatial and temporal contextual information. In this way, it effectively regularizes spatial prediction space within each image and sequential variation space across temporal frames, thereby reducing ambiguous predictions. To verify the effectiveness of exploring these spatial-temporal correlations, we implement the baseline STTran method for comparison purposes, setting the layers of its spatial encoder and temporal decoder to two for fair comparisons. As shown in Table \ref{tab:ablation-study}, the baseline STTran method obtains the R@10 and R@20 values of 77.8\% and 94.1\% on \textit{Pred Cls} and the R@10 and R@20 values of 53.8\% and 63.6\% on \textit{SG Cls}. By incorporating spatial-temporal correlations to regularize training and inference, the STKET boosts the R@10 and R@20 values to 82.6\% and 96.3\% on \textit{Pred Cls} and the R@10 and R@20 values to 57.1\% and 65.3\% on \textit{SG Cls}. Similarly, it consistently outperforms the baseline method on other metrics(i.e., the mR@10 and mR@20 metrics), as shown in Table \ref{tab:ablation-study}.

\begin{table*}[!t]
  \setlength\tabcolsep{5pt}
  \centering
  \caption{Comparison of the baseline STTran method (STTran), the baseline STTran method merely using the spatial encoder (STTran SE), the baseline STTran method merely using the temporal decoder (STTran TD), our framework merely using SKEL (Our SKEL), our framework merely using SKEL without the loss $\mathcal{L}_{spk}$ (Ours SKEL w/o $\mathcal{L}_{spk}$), our framework merely using TKEL (Our TKEL), our framework merely using TKEL without the loss $\mathcal{L}_{tpk}$ (Ours TKEL w/o $\mathcal{L}_{tpk}$), our framework removing STA (Ours w/o STA) and our framework (Ours).}
  \begin{tabular}{c|cccc|cccc}
  \hline
  \hline
  \centering \multirow{2}{*}{Methods} & \multicolumn{4}{c}{Pred Cls} & \multicolumn{4}{c}{SG Cls} \\
  \cline{2-9}
  \centering & R@10 & R@20 & mR@10 & mR@20 & R@10 & R@20 & mR@10 & mR@20 \\
  \hline
  \centering STTran & 77.8 & 94.1 & 15.4 & 19.6 & 53.8 & 63.6 & 10.7 & 13.1 \\
  \centering STTran SE & 75.3 & 91.1 & 13.8 & 18.9 & 51.6 & 62.1 & 9.5 & 12.2 \\
  \centering STTran TD & 76.7 & 93.2 & 14.5 & 19.2 & 53.1 & 63.0 & 10.0 & 12.6 \\
  \hline
  \centering Ours SKEL & 78.0 & 94.3 & 15.5 & 19.8 & 54.2 & 64.1 & 11.3 & 13.5 \\
  \centering Ours SKEL w/o $\mathcal{L}_{spk}$ & 77.3 & 93.9 & 14.9 & 19.3 & 53.6 & 63.7 & 10.5 & 12.8 \\
  \hline
  \centering Ours TKEL & 81.2 & 95.5 & 18.7 & 25.1 & 56.0 & 64.4 & 14.3 & 16.2 \\
  \centering Ours TKEL w/o $\mathcal{L}_{tpk}$ & 80.7 & 95.1 & 18.2 & 24.4 & 55.3 & 63.7 & 13.6 & 15.8 \\
  \hline
  \centering Ours w/o STA & 81.9 & 95.8 & 19.4 & 26.5 & 56.5 & 64.9 & 14.8 & 16.7 \\
  \centering Ours & \textbf{82.6} & \textbf{96.3} & \textbf{20.1} & \textbf{27.3} & \textbf{57.1} & \textbf{65.3} & \textbf{15.4} & \textbf{17.2} \\
  \hline
  \hline
  \end{tabular}
  \vspace{0pt}
  \label{tab:ablation-study}
  \vspace{0pt}
\end{table*}

Further, to emphasize the effectiveness of spatial-temporal knowledge, we conduct qualitative experiments that visualize two relationship transition matrices of different subject-object pairs (i.e., ``man-tv" and ``man-cup") in Figure \ref{fig:entropy}, where a lighter color indicates a higher probability of relationship transition. It's worth noting that transition probabilities vary widely depending on the predicates and object categories involved, e.g., the transition likelihood for common predicates like ``looking at" is much higher than for rare predicates such as ``writing on". This variance within the transition matrix highlights the importance of incorporating spatial-temporal knowledge for effective relationship prediction regularization. Additionally, we evaluate the entropy of spatial-temporal knowledge embeddings generated by our STKET model. Lower entropy values in these embeddings, particularly those for infrequent relationship categories like ``eating", ``writing on", and ``twisting", indicate a significant amount of prior information. This suggests that these embeddings offer substantial prediction regularization, potentially explaining why our STKET model excels over existing leading algorithms in predicting less common relationships.

Since the STKET framework consists of three complementary modules, i.e., the SKEL module, the TKEL module, and the STA module, in the following, we further conduct more ablation experiments to analyze the actual contribution of each module for a more in-depth understanding.

\subsubsection{Analysis of SKEL} To evaluate the actual contribution of the SKEL module, we compare the performance of our STKEL framework merely using this module (namely, ``Ours SKEL") with the performance of the baseline STTran method merely using its spatial encoder (namely, ``STTran SE"). As shown in Table \ref{tab:ablation-study}, the SKEL module improves the R@10 from 75.3\% to 78.0\% and the R@20 from 91.1\% to 94.3\% on \textit{Pred Cls}, with improvements of 2.7\% and 3.2\%, respectively. Similarly, it improves the R@10 from 51.6\% to 54.2\% and the R@20 from 62.1\% to 64.1\% on \textit{SG Cls}, with improvements of 2.6\% and 2.0\%, respectively. It is worth noting that the SKEL module achieves not only performance improvement on the R@10 and R@20 metrics but also on the mR@10 and mR@20 metrics. Specifically, this module obtains an mR@10 improvement of 1.7\% and 1.8\% and an mR@20 improvement of 0.9\% and 1.3\% on \textit{Pred Cls} and \textit{SG Cls}, respectively. These results demonstrate that the spatial co-occurrence correlation within each image can help to regularize spatial prediction space within each image effectively.

In the SKEL module, the loss $\mathcal{L}_{spk}$ helps to learn accurate spatial knowledge embedding, thereby effectively regularizing spatial prediction space within each image. To evaluate the contribution of this loss, we conduct experiments that only utilize the SKEL module without the loss (namely, ``Ours SKEL w/o $\mathcal{L}_{spk}$") for comparison purposes. As presented in Table \ref{tab:ablation-study}, it decreases the performance by 0.7\%/0.4\% on \textit{Pred Cls} with R@10/20 and 0.6\%/0.4\% on \textit{SG Cls} with R@10/20. Similarly, it degrades performance by 0.6\%/0.5\% on \textit{Pred Cls} with mR@10/20 and 0.8\%/0.7\% on \textit{SG Cls} with mR@10/20.

\subsubsection{Analysis of TKEL} To evaluate the actual contribution of the TKEL module, we conduct experiments that remove the SKEL and STA module in STKET while only using this module (namely, ``Our TKEL") and compare it with the baseline STTran method merely using its temporal decoder (namely, ``STTran TD") for comparison purposes. Different from ``STTran SE" which explores the spatial context within the single frame, ``STTran TD" introduces a sliding window to capture the temporal dependencies between frames and thus achieves obvious performance improvement, as presented in Table \ref{tab:ablation-study}. Specifically, it improves the performance by 1.4\%/2.1\% on \textit{Pred Cls} with R@10/20 and 1.5\%/0.9\% on \textit{SG Cls} with R@10/20, which demonstrates the importance of temporal correlations in recognizing dynamic visual relationships. However, ``Our TKEL" performs better than this baseline ``STTran TD", with an R@10/20 improvement of 4.5\% and 4.4\% on \textit{Pred Cls} and an R@10/20 improvement of 2.9\% and 1.4\% on \textit{SG Cls}. Furthermore, ``Our TKEL" consistently outperforms the baseline ``STTran TD" in terms of the mR@10 and mR@20 metrics, i.e., it provides an R@10/20 improvement of 4.2\% and 5.9\% on \textit{Pred Cls}and an mR@10/20 improvement of 4.3\% and 3.6\% on \textit{SG Cls}. These results demonstrate that integrating temporal knowledge can be helpful in regularizing model learning correct temporal correlations, thereby effectively reducing ambiguous predictions.

In the TKEL module, the loss $\mathcal{L}_{tpk}$helps to learn accurate temporal knowledge embedding, thereby effectively regularizing sequential variation space across temporal frames. To evaluate the contribution of this loss, we conduct experiments that merely use TKEL without the loss (namely, ``Ours TKEL w/o $\mathcal{L}_{tpk}$") for comparison purposes. As presented, it decreases the performance by 0.5\%/0.4\% on \textit{Pred Cls} with R@10/20 and 0.7\%/0.7\% on \textit{SG Cls} with R@10/20. Similarly, it degrades performance by 0.5\%/0.7\% on \textit{Pred Cls} with mR@10/20 and 0.7\%/0.4\% on \textit{SG Cls} with mR@10/20.

\begin{figure}[!t] 
  \centering    
  \subfloat[Impact of sliding window in TKEL]{
  \includegraphics[width=0.9\linewidth]{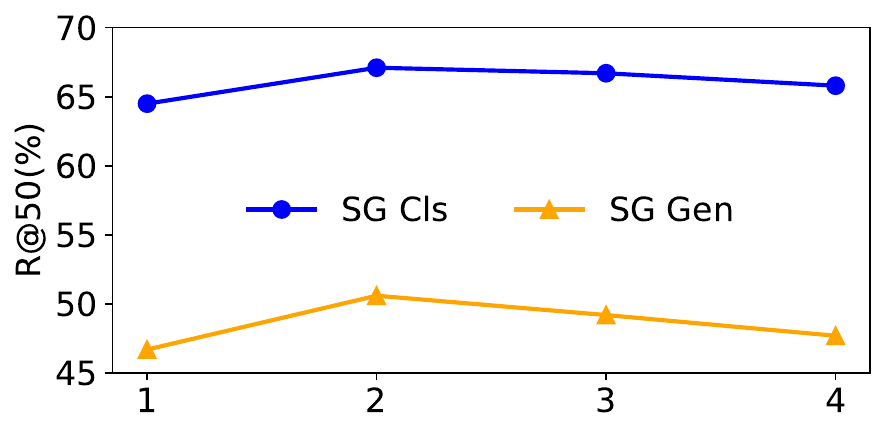}  
  }~     
  
  \subfloat[Impact of sliding window in STA]{
  \includegraphics[width=0.9\linewidth]{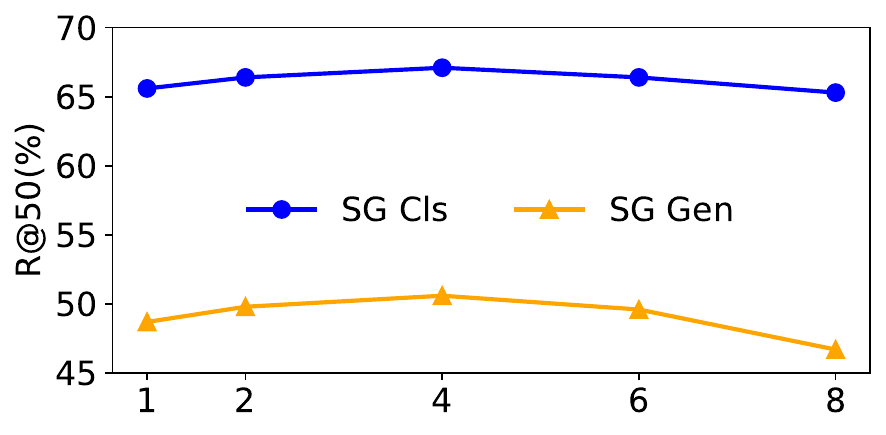} 
  }~
  \caption{Ablation analysis of the frame number of sliding window in (a) TKEL and (b) STA modules. Evaluated on Action Genome \cite{Ji2020AG}.}  
  \label{fig:frame_number}
  \vspace{0pt}
\end{figure}

\subsubsection{Analysis of STA} Considering the sliding window may result in many irrelevant representations, which may easily overwhelm the valuable information, it is difficult for the TKEL module to capture the long-term context of each subject-object pair. Thus, we design the STA module that aggregates spatial- and temporal-embedded representations of the identical subject-object pair across different frames to explore the long-term context. To evaluate its actual contribution, we conduct experiments that remove STA (namely, ``Ours w/o STA"). As shown in Table \ref{tab:ablation-study}, it decreases the performance by 0.7\%/0.5\% on \textit{Pred Cls} with R@10/20 and 0.6\%/0.4\% on \textit{SG Cls} with R@10/20. It also decreases performance by 0.7\%/0.8\% on \textit{Pred Cls} with mR@10/20 and 0.6\%/0.5\% on \textit{SG Cls} with mR@10/20.

\subsubsection{Analysis of Sliding Window} In the TKEL and STA modules, we introduce the sliding window to explore the temporal context contained in previous frames, in which the frame number is a crucial threshold that controls the sequence length. Setting it to a small value may miss temporal correlations, while setting it to a large value may introduce much irrelevant information and result in high computation costs. To figure out the optimal settings, we conduct experiments with different frame numbers. As shown in Figure \ref{fig:frame_number}, introducing more previous frames can significantly improve performance when there is only the current frame (i.e., the frame number is one). Besides, it is worth noting that a large frame number does not always mean better performance because long-term frame sequences contain complicated temporal contexts, which may mislead the model.

\subsection{Limitation}
\begin{figure}[!t] 
  \centering
  \includegraphics[width=0.9\linewidth]{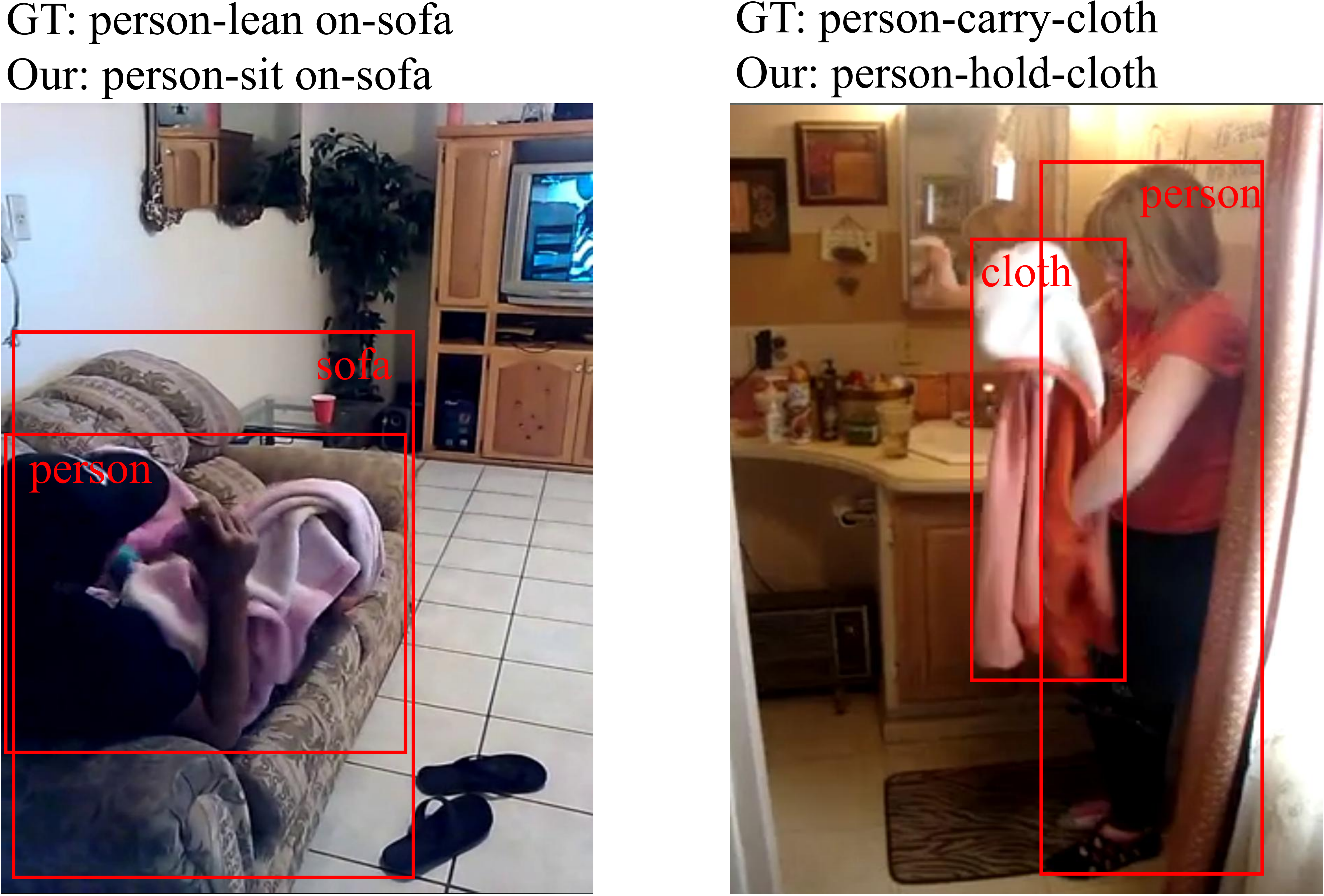}
  \caption{Instances of failure cases resulted from relationship imbalance. For rare subject-predicate-object triplets, the model may predict similar predicates that have higher frequency instead of the ground truth predicate, which occurs less frequently in the training set.}
  \label{fig:failure_case}     
  \vspace{0pt}
\end{figure}

As aforementioned, our STKET framework leverages spatial-temporal prior knowledge to effectively regularize relationship predictions, thereby lessening reliance on training samples and mitigating the imbalance problem in the Action Genome. The significant performance boost within tailed relationship categories demonstrates it, as shown in Figure \ref{fig:recall_comparison}. Nevertheless, in extreme scenarios where subject-predicate-object triplets are exceedingly rare, the STKET framework struggles to provide accurate statistical regularization and thus may predict false predicates. As illustrated in \ref{fig:failure_case}, the bias from skewed long-tail distribution leads to a common but incorrect predicate ``sit on" instead of the rare but correct predicate ``lean on", in the relationship of ``person-lean on-sofa". The same type of misclassification also occur with "person-carry-cloth". For these challenges, we conjecture that integrating spatial-temporal knowledge from various distributions (e.g., ImageNet-VidVRD \cite{Shang2017VidVRD}, Home Action Genome \cite{Rai2021HomeActionGenome}) can provide more accurate and robust regularization. Furthermore, we argue that leveraging the abundant contextual information in web text data could enhance the regularization of such rare relationships.

\section{Conclusion}
In this work, we propose to explore spatial-temporal prior knowledge to facilitate VidSGG via a spatial-temporal knowledge-embedded transformer (STKET) framework. It contains the spatial and temporal knowledge embedded layers that integrate spatial co-occurrence correlations to guide aggregating spatial contextual information and temporal transition correlations to help extract temporal contextual information. In this way, it can, on the one hand, aggregate spatial and temporal information to learn more representative relationship representation and, on the other hand, effectively regularize relationship prediction and thus reduce the dependencies on training samples. Extensive experiments illustrate its superiority over current state-of-the-art algorithms.

\bibliographystyle{IEEEtran}
\bibliography{reference}

% Generated by IEEEtran.bst, version: 1.14 (2015/08/26)
\begin{thebibliography}{10}
\providecommand{\url}[1]{#1}
\csname url@samestyle\endcsname
\providecommand{\newblock}{\relax}
\providecommand{\bibinfo}[2]{#2}
\providecommand{\BIBentrySTDinterwordspacing}{\spaceskip=0pt\relax}
\providecommand{\BIBentryALTinterwordstretchfactor}{4}
\providecommand{\BIBentryALTinterwordspacing}{\spaceskip=\fontdimen2\font plus
\BIBentryALTinterwordstretchfactor\fontdimen3\font minus \fontdimen4\font\relax}
\providecommand{\BIBforeignlanguage}[2]{{%
\expandafter\ifx\csname l@#1\endcsname\relax
\typeout{** WARNING: IEEEtran.bst: No hyphenation pattern has been}%
\typeout{** loaded for the language `#1'. Using the pattern for}%
\typeout{** the default language instead.}%
\else
\language=\csname l@#1\endcsname
\fi
#2}}
\providecommand{\BIBdecl}{\relax}
\BIBdecl

\bibitem{Xu2017IMP}
D.~Xu, Y.~Zhu, C.~B. Choy, and L.~Fei-Fei, ``Scene graph generation by iterative message passing,'' in \emph{Proceedings of the IEEE conference on computer vision and pattern recognition}, 2017, pp. 5410--5419.

\bibitem{Ren2020TNNLS-SGG}
G.~Ren, L.~Ren, Y.~Liao, S.~Liu, B.~Li, J.~Han, and S.~Yan, ``Scene graph generation with hierarchical context,'' \emph{IEEE Transactions on Neural Networks and Learning Systems}, vol.~32, no.~2, pp. 909--915, 2020.

\bibitem{Tao2022TIP-SGG}
L.~Tao, L.~Mi, N.~Li, X.~Cheng, Y.~Hu, and Z.~Chen, ``Predicate correlation learning for scene graph generation,'' \emph{IEEE Transactions on Image Processing}, vol.~31, pp. 4173--4185, 2022.

\bibitem{Tu2019TIP-VideoActionRecognition}
Z.~Tu, H.~Li, D.~Zhang, J.~Dauwels, B.~Li, and J.~Yuan, ``Action-stage emphasized spatiotemporal vlad for video action recognition,'' \emph{IEEE Transactions on Image Processing}, vol.~28, no.~6, pp. 2799--2812, 2019.

\bibitem{Han2022TAN}
T.~Han, W.~Xie, and A.~Zisserman, ``Temporal alignment networks for long-term video,'' in \emph{Proceedings of the IEEE/CVF Conference on Computer Vision and Pattern Recognition}, 2022, pp. 2906--2916.

\bibitem{Zhou2023TIP-VideoActionRecognition}
Z.~Zhou, C.~Ding, J.~Li, E.~Mohammadi, G.~Liu, Y.~Yang, and Q.~M.~J. Wu, ``Sequential order-aware coding-based robust subspace clustering for human action recognition in untrimmed videos,'' \emph{IEEE Transactions on Image Processing}, vol.~32, pp. 13--28, 2023.

\bibitem{Tang2021TCSVT-VideoSegmentation}
Z.~Tang, Y.~Liao, S.~Liu, G.~Li, X.~Jin, H.~Jiang, Q.~Yu, and D.~Xu, ``Human-centric spatio-temporal video grounding with visual transformers,'' \emph{IEEE Transactions on Circuits and Systems for Video Technology}, vol.~32, no.~12, pp. 8238--8249, 2021.

\bibitem{Ding2022BCA}
Z.~Ding, T.~Hui, J.~Huang, X.~Wei, J.~Han, and S.~Liu, ``Language-bridged spatial-temporal interaction for referring video object segmentation,'' in \emph{Proceedings of the IEEE/CVF Conference on Computer Vision and Pattern Recognition}, 2022, pp. 4964--4973.

\bibitem{Hui2023TPAMI-CMAM}
T.~Hui, S.~Liu, Z.~Ding, S.~Huang, G.~Li, W.~Wang, L.~Liu, and J.~Han, ``Language-aware spatial-temporal collaboration for referring video segmentation,'' \emph{IEEE Transactions on Pattern Analysis and Machine Intelligence}, 2023.

\bibitem{Nishimura2021VPC}
T.~Nishimura, A.~Hashimoto, Y.~Ushiku, H.~Kameko, and S.~Mori, ``State-aware video procedural captioning,'' in \emph{Proceedings of the 29th ACM International Conference on Multimedia}, 2021, pp. 1766--1774.

\bibitem{Huang2021SemanticTag}
Y.~Huang, H.~Xue, J.~Chen, H.~Ma, and H.~Ma, ``Semantic tag augmented xlanv model for video captioning,'' in \emph{Proceedings of the 29th ACM International Conference on Multimedia}, 2021, pp. 4818--4822.

\bibitem{Wang2022TIP-VideoCaptioning}
H.~Wang, G.~Lin, S.~C.~H. Hoi, and C.~Miao, ``Cross-modal graph with meta concepts for video captioning,'' \emph{IEEE Transactions on Image Processing}, vol.~31, pp. 5150--5162, 2022.

\bibitem{Hua2022TIP-VideoCaptioning}
X.~Hua, X.~Wang, T.~Rui, F.~Shao, and D.~Wang, ``Adversarial reinforcement learning with object-scene relational graph for video captioning,'' \emph{IEEE Transactions on Image Processing}, vol.~31, pp. 2004--2016, 2022.

\bibitem{Yang2020BertVideoQA}
Z.~Yang, N.~Garcia, C.~Chu, M.~Otani, Y.~Nakashima, and H.~Takemura, ``Bert representations for video question answering,'' in \emph{Proceedings of the IEEE/CVF Winter Conference on Applications of Computer Vision}, 2020, pp. 1556--1565.

\bibitem{Zeng2022TIP-VideoQA}
P.~Zeng, H.~Zhang, L.~Gao, J.~Song, and H.~T. Shen, ``Video question answering with prior knowledge and object-sensitive learning,'' \emph{IEEE Transactions on Image Processing}, vol.~31, pp. 5936--5948, 2022.

\bibitem{Gao2022TIP-VideoQA}
L.~Gao, Y.~Lei, P.~Zeng, J.~Song, M.~Wang, and H.~T. Shen, ``Hierarchical representation network with auxiliary tasks for video captioning and video question answering,'' \emph{IEEE Transactions on Image Processing}, vol.~31, pp. 202--215, 2022.

\bibitem{Liu2022TIP-VideoQA}
Y.~Liu, X.~Zhang, F.~Huang, B.~Zhang, and Z.~Li, ``Cross-attentional spatio-temporal semantic graph networks for video question answering,'' \emph{IEEE Transactions on Image Processing}, vol.~31, pp. 1684--1696, 2022.

\bibitem{Liu2019SGGsurvey}
D.~Liu, M.~Bober, and J.~Kittler, ``Visual semantic information pursuit: A survey,'' \emph{IEEE transactions on pattern analysis and machine intelligence}, vol.~43, no.~4, pp. 1404--1422, 2019.

\bibitem{Xu2020SGGsurvey}
P.~Xu, X.~Chang, L.~Guo, P.-Y. Huang, X.~Chen, and A.~G. Hauptmann, ``A survey of scene graph: Generation and application,'' \emph{IEEE Trans. Neural Netw. Learn. Syst}, vol.~1, 2020.

\bibitem{Chang2021SGGsurvey}
X.~Chang, P.~Ren, P.~Xu, Z.~Li, X.~Chen, and A.~Hauptmann, ``A comprehensive survey of scene graphs: Generation and application,'' \emph{IEEE Transactions on Pattern Analysis and Machine Intelligence}, vol.~45, no.~1, pp. 1--26, 2021.

\bibitem{Shang2017VidVRD}
X.~Shang, T.~Ren, J.~Guo, H.~Zhang, and T.-S. Chua, ``Video visual relation detection,'' in \emph{Proceedings of the 25th ACM international conference on Multimedia}, 2017, pp. 1300--1308.

\bibitem{Qian2019VRD-GCN}
X.~Qian, Y.~Zhuang, Y.~Li, S.~Xiao, S.~Pu, and J.~Xiao, ``Video relation detection with spatio-temporal graph,'' in \emph{Proceedings of the 27th ACM International Conference on Multimedia}, 2019, pp. 84--93.

\bibitem{Zheng2022VRDFormer}
S.~Zheng, S.~Chen, and Q.~Jin, ``Vrdformer: End-to-end video visual relation detection with transformers,'' in \emph{Proceedings of the IEEE/CVF Conference on Computer Vision and Pattern Recognition}, 2022, pp. 18\,836--18\,846.

\bibitem{Li2022APT}
Y.~Li, X.~Yang, and C.~Xu, ``Dynamic scene graph generation via anticipatory pre-training,'' in \emph{Proceedings of the IEEE/CVF Conference on Computer Vision and Pattern Recognition}, 2022, pp. 13\,874--13\,883.

\bibitem{Cong2021STTran}
Y.~Cong, W.~Liao, H.~Ackermann, B.~Rosenhahn, and M.~Y. Yang, ``Spatial-temporal transformer for dynamic scene graph generation,'' in \emph{Proceedings of the IEEE/CVF International Conference on Computer Vision}, 2021, pp. 16\,372--16\,382.

\bibitem{Ji2020AG}
J.~Ji, R.~Krishna, L.~Fei-Fei, and J.~C. Niebles, ``Action genome: Actions as compositions of spatio-temporal scene graphs,'' in \emph{Proceedings of the IEEE/CVF Conference on Computer Vision and Pattern Recognition}, 2020, pp. 10\,236--10\,247.

\bibitem{Zhang2019RelDN}
J.~Zhang, K.~J. Shih, A.~Elgammal, A.~Tao, and B.~Catanzaro, ``Graphical contrastive losses for scene graph parsing,'' in \emph{Proceedings of the IEEE/CVF Conference on Computer Vision and Pattern Recognition}, 2019, pp. 11\,535--11\,543.

\bibitem{Lin2020GPS-Net}
X.~Lin, C.~Ding, J.~Zeng, and D.~Tao, ``Gps-net: Graph property sensing network for scene graph generation,'' in \emph{Proceedings of the IEEE/CVF Conference on Computer Vision and Pattern Recognition}, 2020, pp. 3746--3753.

\bibitem{Ren2020HCNet}
G.~Ren, L.~Ren, Y.~Liao, S.~Liu, B.~Li, J.~Han, and S.~Yan, ``Scene graph generation with hierarchical context,'' \emph{IEEE Transactions on Neural Networks and Learning Systems}, vol.~32, no.~2, pp. 909--915, 2020.

\bibitem{Lu2021Seq2Seq}
Y.~Lu, H.~Rai, J.~Chang, B.~Knyazev, G.~Yu, S.~Shekhar, G.~W. Taylor, and M.~Volkovs, ``Context-aware scene graph generation with seq2seq transformers,'' in \emph{Proceedings of the IEEE/CVF International Conference on Computer Vision}, 2021, pp. 15\,931--15\,941.

\bibitem{Lu2016VRD}
C.~Lu, R.~Krishna, M.~Bernstein, and L.~Fei-Fei, ``Visual relationship detection with language priors,'' in \emph{European conference on computer vision}.\hskip 1em plus 0.5em minus 0.4em\relax Springer, 2016, pp. 852--869.

\bibitem{Li2017MSDN}
Y.~Li, W.~Ouyang, B.~Zhou, K.~Wang, and X.~Wang, ``Scene graph generation from objects, phrases and region captions,'' in \emph{Proceedings of the IEEE international conference on computer vision}, 2017, pp. 1261--1270.

\bibitem{Yang2018aGCN}
J.~Yang, J.~Lu, S.~Lee, D.~Batra, and D.~Parikh, ``Graph r-cnn for scene graph generation,'' in \emph{Proceedings of the European conference on computer vision (ECCV)}, 2018, pp. 670--685.

\bibitem{Zellers2018Motif}
R.~Zellers, M.~Yatskar, S.~Thomson, and Y.~Choi, ``Neural motifs: Scene graph parsing with global context,'' in \emph{Proceedings of the IEEE conference on computer vision and pattern recognition}, 2018, pp. 5831--5840.

\bibitem{Kipf2016GCN}
T.~N. Kipf and M.~Welling, ``Semi-supervised classification with graph convolutional networks,'' \emph{arXiv preprint arXiv:1609.02907}, 2016.

\bibitem{Chen2019KERN}
T.~Chen, W.~Yu, R.~Chen, and L.~Lin, ``Knowledge-embedded routing network for scene graph generation,'' in \emph{Proceedings of the IEEE/CVF Conference on Computer Vision and Pattern Recognition}, 2019, pp. 6163--6171.

\bibitem{Tang2019VCTree}
K.~Tang, H.~Zhang, B.~Wu, W.~Luo, and W.~Liu, ``Learning to compose dynamic tree structures for visual contexts,'' in \emph{Proceedings of the IEEE/CVF conference on computer vision and pattern recognition}, 2019, pp. 6619--6628.

\bibitem{Vaswani2017Transformer}
A.~Vaswani, N.~Shazeer, N.~Parmar, J.~Uszkoreit, L.~Jones, A.~N. Gomez, {\L}.~Kaiser, and I.~Polosukhin, ``Attention is all you need,'' \emph{Advances in neural information processing systems}, vol.~30, 2017.

\bibitem{Cong2023Reltr}
Y.~Cong, M.~Y. Yang, and B.~Rosenhahn, ``Reltr: Relation transformer for scene graph generation,'' \emph{IEEE Transactions on Pattern Analysis and Machine Intelligence}, 2023.

\bibitem{Kundu2023IS-GGT}
S.~Kundu and S.~N. Aakur, ``Is-ggt: Iterative scene graph generation with generative transformers,'' in \emph{Proceedings of the IEEE/CVF Conference on Computer Vision and Pattern Recognition}, 2023, pp. 6292--6301.

\bibitem{Tsai2019GSTEG}
Y.-H.~H. Tsai, S.~Divvala, L.-P. Morency, R.~Salakhutdinov, and A.~Farhadi, ``Video relationship reasoning using gated spatio-temporal energy graph,'' in \emph{Proceedings of the IEEE/CVF Conference on Computer Vision and Pattern Recognition}, 2019, pp. 10\,424--10\,433.

\bibitem{Liu2020BeyondShortSnippet}
C.~Liu, Y.~Jin, K.~Xu, G.~Gong, and Y.~Mu, ``Beyond short-term snippet: Video relation detection with spatio-temporal global context,'' in \emph{Proceedings of the IEEE/CVF conference on computer vision and pattern recognition}, 2020, pp. 10\,840--10\,849.

\bibitem{Teng2021TRACE}
Y.~Teng, L.~Wang, Z.~Li, and G.~Wu, ``Target adaptive context aggregation for video scene graph generation,'' in \emph{Proceedings of the IEEE/CVF International Conference on Computer Vision}, 2021, pp. 13\,688--13\,697.

\bibitem{Wang2022TPI}
S.~Wang, L.~Gao, X.~Lyu, Y.~Guo, P.~Zeng, and J.~Song, ``Dynamic scene graph generation via temporal prior inference,'' in \emph{Proceedings of the 30th ACM International Conference on Multimedia}, 2022, pp. 5793--5801.

\bibitem{Kumar2023FS-SGG}
Y.~Kumar and A.~Mishra, ``Few-shot referring relationships in videos,'' in \emph{Proceedings of the IEEE/CVF Conference on Computer Vision and Pattern Recognition}, 2023, pp. 2289--2298.

\bibitem{Wang2020SAOA}
X.~Wang, L.~Zhu, Y.~Wu, and Y.~Yang, ``Symbiotic attention for egocentric action recognition with object-centric alignment,'' \emph{IEEE transactions on pattern analysis and machine intelligence}, 2020.

\bibitem{He2016ResNet}
K.~He, X.~Zhang, S.~Ren, and J.~Sun, ``Deep residual learning for image recognition,'' in \emph{Proceedings of the IEEE conference on computer vision and pattern recognition}, 2016, pp. 770--778.

\bibitem{Li2022MsKAT}
H.~Li, C.~Li, A.~Zheng, J.~Tang, and B.~Luo, ``Mskat: Multi-scale knowledge-aware transformer for vehicle re-identification,'' \emph{IEEE Transactions on Intelligent Transportation Systems}, vol.~23, no.~10, pp. 19\,557--19\,568, 2022.

\bibitem{Wang2023Restoreformer++}
Z.~Wang, J.~Zhang, T.~Chen, W.~Wang, and P.~Luo, ``Restoreformer++: Towards real-world blind face restoration from undegraded key-value pairs,'' \emph{IEEE Transactions on Pattern Analysis and Machine Intelligence}, 2023.

\bibitem{Chen2022SST}
T.~Chen, T.~Pu, H.~Wu, Y.~Xie, and L.~Lin, ``Structured semantic transfer for multi-label recognition with partial labels,'' in \emph{Proceedings of the AAAI conference on artificial intelligence}, vol.~36, no.~1, 2022, pp. 339--346.

\bibitem{Pu2022SARB}
T.~Pu, T.~Chen, H.~Wu, and L.~Lin, ``Semantic-aware representation blending for multi-label image recognition with partial labels,'' in \emph{Proceedings of the AAAI conference on artificial intelligence}, vol.~36, no.~2, 2022, pp. 2091--2098.

\bibitem{Xie2020CD-FER}
Y.~Xie, T.~Chen, T.~Pu, H.~Wu, and L.~Lin, ``Adversarial graph representation adaptation for cross-domain facial expression recognition,'' in \emph{Proceedings of the 28th ACM international conference on Multimedia}, 2020.

\bibitem{Peng2019KTN}
Z.~Peng, Z.~Li, J.~Zhang, Y.~Li, G.-J. Qi, and J.~Tang, ``Few-shot image recognition with knowledge transfer,'' in \emph{Proceedings of the IEEE/CVF International Conference on Computer Vision (ICCV)}, October 2019.

\bibitem{Chen2022KGGR}
T.~Chen, L.~Lin, R.~Chen, X.~Hui, and H.~Wu, ``Knowledge-guided multi-label few-shot learning for general image recognition,'' \emph{IEEE Transactions on Pattern Analysis and Machine Intelligence}, vol.~44, no.~3, pp. 1371--1384, 2022.

\bibitem{Pu2021AUE-CRL}
T.~Pu, T.~Chen, Y.~Xie, H.~Wu, and L.~Lin, ``Au-expression knowledge constrained representation learning for facial expression recognition,'' in \emph{2021 IEEE international conference on robotics and automation (ICRA)}.\hskip 1em plus 0.5em minus 0.4em\relax IEEE, 2021, pp. 11\,154--11\,161.

\bibitem{Chen2022CD-FER}
T.~Chen, T.~Pu, H.~Wu, Y.~Xie, L.~Liu, and L.~Lin, ``Cross-domain facial expression recognition: A unified evaluation benchmark and adversarial graph learning,'' \emph{IEEE Transactions on Pattern Analysis and Machine Intelligence}, vol.~44, no.~12, pp. 9887--9903, 2022.

\bibitem{Yang2019ScenePrior}
W.~Yang, X.~Wang, A.~Farhadi, A.~Gupta, and R.~Mottaghi, ``Visual semantic navigation using scene priors,'' in \emph{Proceedings of International Conference on Learning Representations (ICLR)}, 2019.

\bibitem{Ren2015FasterRCNN}
S.~Ren, K.~He, R.~Girshick, and J.~Sun, ``Faster r-cnn: Towards real-time object detection with region proposal networks,'' \emph{Advances in neural information processing systems}, vol.~28, 2015.

\bibitem{He2017MaskRCNN}
K.~He, G.~Gkioxari, P.~Doll{\'a}r, and R.~Girshick, ``Mask r-cnn,'' in \emph{Proceedings of the IEEE international conference on computer vision}, 2017, pp. 2961--2969.

\bibitem{Vandenbroucke2016Prior}
A.~R. Vandenbroucke, J.~Fahrenfort, J.~Meuwese, H.~Scholte, and V.~Lamme, ``Prior knowledge about objects determines neural color representation in human visual cortex,'' \emph{Cerebral cortex}, vol.~26, no.~4, pp. 1401--1408, 2016.

\bibitem{Lee2018ML-ZSL}
C.-W. Lee, W.~Fang, C.-K. Yeh, and Y.-C.~F. Wang, ``Multi-label zero-shot learning with structured knowledge graphs,'' in \emph{Proceedings of the IEEE conference on computer vision and pattern recognition}, 2018, pp. 1576--1585.

\bibitem{Loshchilov2017Adam}
I.~Loshchilov and F.~Hutter, ``Decoupled weight decay regularization,'' \emph{arXiv preprint arXiv:1711.05101}, 2017.

\bibitem{Paszke2019Pytorch}
A.~Paszke, S.~Gross, F.~Massa, A.~Lerer, J.~Bradbury, G.~Chanan, T.~Killeen, Z.~Lin, N.~Gimelshein, L.~Antiga \emph{et~al.}, ``Pytorch: An imperative style, high-performance deep learning library,'' \emph{Advances in neural information processing systems}, vol.~32, 2019.

\bibitem{Rai2021HomeActionGenome}
N.~Rai, H.~Chen, J.~Ji, R.~Desai, K.~Kozuka, S.~Ishizaka, E.~Adeli, and J.~C. Niebles, ``Home action genome: Cooperative compositional action understanding,'' in \emph{Proceedings of the IEEE/CVF Conference on Computer Vision and Pattern Recognition}, 2021, pp. 11\,184--11\,193.

\end{thebibliography}

%\newpage

\begin{IEEEbiography}[{\includegraphics[width=1in,height=1.25in,clip,keepaspectratio]{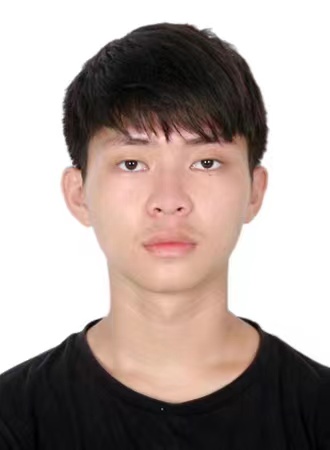}}]{Tao Pu} received a B.E. degree from the School of Computer Science and Engineering, Sun Yat-sen University, Guangzhou, China, in 2020, where he is currently pursuing a Ph.D. degree in computer science. He has authored and coauthored approximately 10 papers published in top-tier academic journals and conferences, including T-PAMI, AAAI, ACM MM, etc.\end{IEEEbiography}

\begin{IEEEbiography}[{\includegraphics[width=1in,height=1.25in,clip,keepaspectratio]{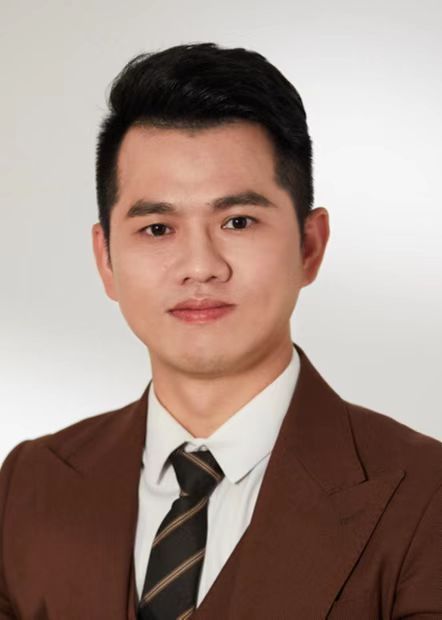}}]{Tianshui Chen} received a Ph.D. degree in computer science at the School of Data and Computer Science Sun Yat-sen University, Guangzhou, China, in 2018. Prior to earning his Ph.D, he received a B.E. degree from the School of Information and Science Technology in 2013. He is currently an associated professor in the Guangdong University of Technology. His current research interests include computer vision and machine learning. He has authored and coauthored approximately 40 papers published in top-tier academic journals and conferences, including T-PAMI, T-NNLS, T-IP, T-MM, CVPR, ICCV, AAAI, IJCAI, ACM MM, etc. He has served as a reviewer for numerous academic journals and conferences. He was the recipient of the Best Paper Diamond Award at IEEE ICME 2017. \end{IEEEbiography}

\begin{IEEEbiography}[{\includegraphics[width=1in,height=1.25in,clip,keepaspectratio]{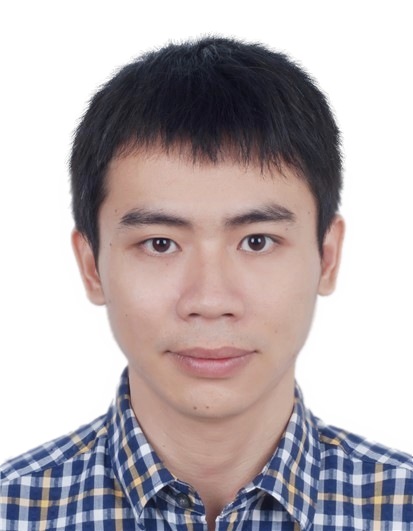}}]{Hefeng Wu} received a B.S. degree in computer science and technology and a Ph.D. degree in computer application technology from Sun Yat-sen University, Guangzhou, China. He is currently a research professor at the School of Computer Science and Engineering, Sun Yat-sen University, China. His research interests include computer vision, multimedia, and machine learning. He has published works in and served as reviewers for many top-tier academic journals and conferences, including T-PAMI, T-IP, T-MM, CVPR, ICCV, AAAI, ACM MM, etc. \end{IEEEbiography}

\begin{IEEEbiography}[{\includegraphics[width=1in,height=1.25in,clip,keepaspectratio]{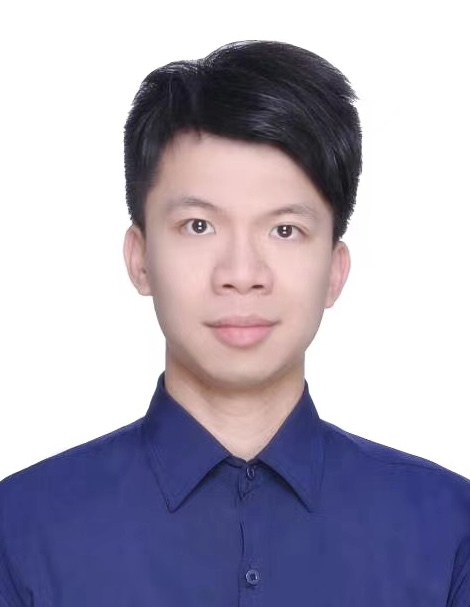}}]{Yongyi Lu} is currently an Associate Professor at Guangdong University of Technology (GDUT). He received his Ph.D. in Computer Science and Engineering at Hong Kong University of Science and Technology (HKUST) in 2018. From 2019-2022, he was a Postdoctoral Fellow in Prof. Alan Yuille’s group at Johns Hopkins University. He received his B.E. and Master both at Sun Yat-Sen University, Guangzhou, China in 2011 and 2014. His research focuses on computer vision, medical image analysis and the interdisciplinary fields of them. He has published over 30 peer-reviewed conference/journal articles such as CVPR, ICCV, ECCV, NeurIPS, MICCAI, BMVC and T-IP, with a total citation of over 2500. He has won the 4th place in ImageNet DET challenge in 2014 and 5th place in ImageNet VID challenge in 2015. He serves as a reviewer of conferences/journals including IJCV, T-IP, T-MM, T-CSVT, Nature Machine Intelligence, Neurocomputing, CVPR, ICCV, ECCV, NeurIPS, ICLR and BMVC. He is a member of IEEE. \end{IEEEbiography}

\begin{IEEEbiography}[{\includegraphics[width=1in,clip]{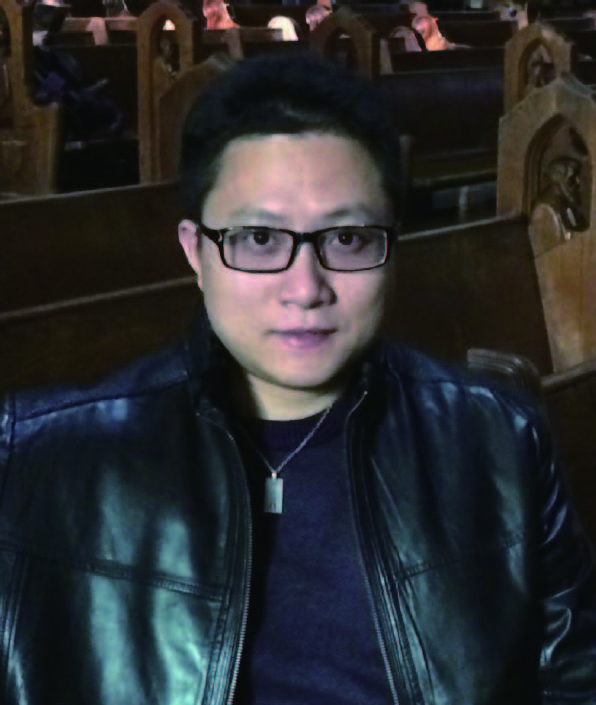}}]{Liang Lin} (Fellow, IEEE) is a full professor at Sun Yat-sen University. From 2008 to 2010, he was a postdoctoral fellow at the University of California, Los Angeles. From 2016--2018, he led the SenseTime R\&D teams to develop cutting-edge and deliverable solutions for computer vision, data analysis and mining, and intelligent robotic systems. He has authored and coauthored more than 100 papers in top-tier academic journals and conferences (e.g., 15 papers in TPAMI and IJCV and 60+ papers in CVPR, ICCV, NIPS, and IJCAI). He has served as an associate editor of IEEE Trans. Human-Machine Systems, The Visual Computer, and Neurocomputing and as an area/session chair for numerous conferences, such as CVPR, ICME, ACCV, and ICMR. He was the recipient of the Annual Best Paper Award by Pattern Recognition (Elsevier) in 2018, the Best Paper Diamond Award at IEEE ICME 2017, the Best Paper Runner-Up Award at ACM NPAR 2010, Google Faculty Award in 2012, the Best Student Paper Award at IEEE ICME 2014, and the Hong Kong Scholars Award in 2014. He is a Fellow of IEEE, IAPR, and IET. \end{IEEEbiography}

\end{document}